\documentclass[journal]{IEEEtran}
\usepackage{amsfonts}
\usepackage{cite}
\usepackage{algorithmicx}
\usepackage{algorithm}
\usepackage{algpseudocode}
\usepackage{graphicx}
\usepackage{subfigure}
\usepackage{amsmath}
\usepackage{multirow}
\usepackage{array}
\usepackage{xcolor}
\usepackage{bm}
\usepackage{mathtools}
\usepackage{amssymb}
\usepackage[hyphens]{url}

\newcommand{\myimgsize}{0.7}
\newcommand{\myimgsizer}{1.2}

\begin{document}
\title{A Spectral Diffusion Prior for Hyperspectral Image Super-Resolution}

\author{Jianjun~Liu,~\IEEEmembership{Member,~IEEE,}
        Zebin~Wu,~\IEEEmembership{Senior~Member,~IEEE,}
        and Liang~Xiao,~\IEEEmembership{Member,~IEEE}
\thanks{Jianjun Liu is with the Jiangsu Provincial Engineering Laboratory for Pattern Recognition and Computational Intelligence, Jiangnan University, Wuxi, China
(Email: liuofficial@163.com).}
\thanks{Zebin Wu and Liang Xiao are with the School of Computer Science,
Nanjing University of Science and Technology, Nanjing, China
(Email: zebin.wu@gmail.com, xiaoliang@mail.njust.edu.cn).}
}

\maketitle

\begin{abstract}
Fusion-based hyperspectral image (HSI) super-resolution aims to produce a high-spatial-resolution HSI by fusing a low-spatial-resolution HSI and a high-spatial-resolution multispectral image. Such a HSI super-resolution process can be modeled as an inverse problem, where the prior knowledge is essential for obtaining the desired solution. Motivated by the success of diffusion models, we propose a novel spectral diffusion prior for fusion-based HSI super-resolution. Specifically, we first investigate the spectrum generation problem and design a spectral diffusion model to model the spectral data distribution. Then, in the framework of maximum a posteriori, we keep the transition information between every two neighboring states during the reverse generative process, and thereby embed the knowledge of trained spectral diffusion model into the fusion problem in the form of a regularization term. At last, we treat each generation step of the final optimization problem as its subproblem, and employ the Adam to solve these subproblems in a reverse sequence. Experimental results conducted on both synthetic and real datasets demonstrate the effectiveness of the proposed approach. The code of the proposed approach will be available on \url{https://github.com/liuofficial/SDP}.
\end{abstract}


\IEEEpeerreviewmaketitle

\section{Introduction}
\IEEEPARstart{H}{yperspectral} images (HSIs) can be treated as a kind of 3-D images containing one-dimensional spectral and two-dimensional spatial information. Different from the general multi-channel images, such as RGB images, HSIs are imaged at hundreds of contiguous and narrow spectral bands, and the spectral range spans the visible to infrared spectrum. Such a high spectral resolution of HSIs makes it possible to accurately identify substances, thus facilitating their practical applications in many fields \cite{Jose2013}, such as agriculture, mineralogy, and military. Conversely, the spatial resolution of HSIs is usually relatively low due to the limitation of imaging sensors \cite{loncan2015hyperspectral,yokoya2017hyperspectral,Dian2021RecentAA,Li2022DeepLI}, which hinders their further development. There are two categories of super-resolution methods developed to obtain high-spatial-resolution HSIs (HR-HSIs). One is the fusion-based HSI super-resolution that instead records a low-spatial-resolution HSI (LR-HSI) and a conventional high-spatial-resolution image to generate the target HR-HSI. The other is the single HSI super-resolution that performs the task without the auxiliary images. Compared with the HSIs, the conventional images such as panchromatic and multispectral images (MSIs) can achieve higher spatial resolution with loss of much spectral information. Benefiting from the complementary information between the two observed images, the fusion-based HSI super-resolution has become a satisfactory solution.

The fusion-based HSI super-resolution mainly refers to the fusion of a LR-HSI and a high-spatial-resolution MSI (HR-MSI) \cite{yokoya2017hyperspectral,Dian2021RecentAA,Li2022DeepLI}, and the fusion of a LR-HSI and a high-spatial-resolution panchromatic image \cite{loncan2015hyperspectral} can be treated as its a special case. This fusion task is similar to the Pansharpening that generates a HR-MSI by fusing a low-spatial-resolution MSI and a high-spatial-resolution
panchromatic image \cite{meng2019review,Vivone2021A}, but it is relatively more complicated due to the need to maintain rich spectral information. Typically, such fusion-based approaches can be roughly classified into four categories: component substitution \cite{tu2001new}, multiresolution analysis \cite{nencini2007remote}, model-based approaches
\cite{yokoya2011coupled,lanaras2015hyperspectral,simoes2015convex,wei2015hyperspectral,liu2020truncated,dong2016hyperspectral,veganzones2016hyperspectral,han2020hyperspectral,li2018fusing,kanatsoulis2018hyperspectral,Xu2020Hyperspectral,Chen2022Hyperspectral,Zhang2018Exploiting,Zhang2017Multispectral,Xue2021Spatial,dian2019hyperspectral}
and deep learning-based approaches
\cite{scarpa2018target,yuan2018a,zhang2019pan,Jiang2020LearningSP,Zhang2021SSR,Dong2021Remote,yang2017pannet,xie2019hyperspectral,fu2021deep,Li2020Hyperspectral,Dong2021GenerativeDN,Bandara2022HyperTransformerAT,wang2023,xu2023}.
For the fusion-based HSI super-resolution, the model-based and deep learning-based approaches have always been the focus of research.

The model-based approaches treat HSI super-resolution as an inverse problem by modeling the spectral and spatial degradation processes. They focus on designing effective fidelity terms to characterize the degradation process and exploiting efficient prior information to constrain the range of target solutions. Various fidelity and regularization/constraint terms have been proposed when designing the model-based approaches, and have shown good performance in HSI super-resolution
\cite{yokoya2011coupled,lanaras2015hyperspectral,simoes2015convex,wei2015hyperspectral,liu2020truncated,dong2016hyperspectral,veganzones2016hyperspectral,han2020hyperspectral,li2018fusing,kanatsoulis2018hyperspectral,Xu2020Hyperspectral,Chen2022Hyperspectral,Zhang2018Exploiting,Zhang2017Multispectral,Xue2021Spatial,dian2019hyperspectral}. The deep learning-based approaches mainly consider building a nonlinear mapping between the observed images and the target image. They typically require large amounts of labelled training data, and focus on designing complete networks to fit the mapping. There are plenty of excellent networks that have been proposed for HSI super-resolution
\cite{scarpa2018target,yuan2018a,zhang2019pan,Jiang2020LearningSP,Zhang2021SSR,Dong2021Remote,yang2017pannet,xie2019hyperspectral,fu2021deep,Li2020Hyperspectral,Dong2021GenerativeDN,Bandara2022HyperTransformerAT,wang2023,xu2023}.
Recently, the complementarity between these two categories of approaches has attracted the attention of researchers \cite{Zhang2021DeepBH,dian2021regularizing,MHF,Dong2021Model,wang2023,xu2023,dong2023,uSDN,FusionNet,MIAE,Gao2023,guo2023,Dian2023}. For example, in the model-based approaches, some researchers employ deep networks as priors to enforce the desired results \cite{Zhang2021DeepBH,dian2021regularizing}. In the deep learning-based approaches, some researchers draw on degradation models to construct elaborate networks or implement unsupervised and zero-shot super-resolution \cite{MHF,Dong2021Model,wang2023,xu2023,dong2023,uSDN,FusionNet,MIAE,Gao2023,guo2023,Dian2023}.

In this paper, we are desired to develop a novel deep prior for the model-based HSI super-resolution. Diffusion models belong to the popular ones at the present, and they have achieved excellent results in many domains such as image and video due to their high generation quality \cite{Luo2022UnderstandingDM,Yang2022DiffusionMA}. Motivated by this, we consider the usage of diffusion models to generate high quality spectra, and thereby exploit a spectral diffusion prior (SDP) for fusion-based HSI super-resolution. Specifically, to apply diffusion models to HSIs, we turn to design a new spectral diffusion model by taking into account the limitation of hyperspectral scenarios. Different from existing image-based or video-based diffusion models, the spectrum diffusion model deals with the spectral data and adopts a multilayer perceptron (MLP) network as its denoising network. To employ the spectral diffusion model as a prior, we assume the target spectra of HR-HSI follow a given spectral data distribution and then keep the transition of every two neighboring Markov states in the inverse generative process of the spectral diffusion model. In the framework of maximum a posteriori, the information preservation of the spectral generative process is eventually depicted by a regularization term. The final optimization problem consists of several subproblems corresponding to timesteps. To solve it, we treat the target HR-HSI as trainable parameters, and adopt the Adam to update the parameters by following the spectral reverse generative sequence.

Compared with the existing HSI super-resolution approaches, some of the innovative characteristics of the proposed SDP are highlighted as follows.
\begin{enumerate}
  \item A spectral diffusion model that models the spectral data distribution of HSIs is proposed to generate high quality spectra. Since it is a pixel-wise model, the requirement of training data and computational resource is greatly reduced.
  \item A spectral diffusion prior is proposed for HSI super-resolution. This approach integrates the fusion process and the spectral generative process into a unified optimization problem. Thus, it can effectively transfer the knowledge of the hyperspectral data at hand to guide the fusion process.
  \item We show that the spectrum generation quality is positively correlated with the fusion accuracy. Thus, this approach opens a wide field for future developments of diffusion models in hyperspectral scenarios.
\end{enumerate}

The remainder of this paper is organized as follows. Section \ref{sec_bk} briefly introduces fusion-based HSI super-resolution and diffusion models and reviews their related works. In Section \ref{sec_prop}, we first introduce the spectral diffusion model and then present the proposed SDP as well as its optimization algorithm. Section \ref{sec_ex} demonstrates the effectiveness of the proposed approach by conducting several experiments. Section \ref{sec_con} concludes this work with some remarks.

\section{Background and Related Works}
\label{sec_bk}
In this section, we briefly review the background of fusion-based HSI super-resolution and diffusion models, as well as their related works.

\subsection{HSI super-resolution}
For fusion-based HSI super-resolution, its goal is to make full use of the complementary information of LR-HSI and HR-MSI to generate the desired HR-HSI. Mathematically, the two observed images can be modeled as spatially degraded and spectrally degraded versions of the target image:
\begin{eqnarray}
  {\bf Y} &=& {\bf X} {\bf BD} + {\bf E}_y, \label{eq_hsi}\\
  {\bf Z} &=& {\bf R} {\bf X} + {\bf E}_z, \label{eq_msi}
\end{eqnarray}
where ${\bf X} \in \mathbb{R}^{N_B \times N_H N_W}$ is the target HR-HSI, with $N_B$, $N_H$ and $N_W$ as its spectral band, height and width, respectively. ${\bf Y} \in \mathbb{R}^{N_B \times N_h N_w}$ and ${\bf Z} \in \mathbb{R}^{N_b \times N_H N_W}$ are the observed LR-HSI and HR-MSI with $N_b$ ($N_b < N_B$), $N_h$ ($N_h < N_H$) and $N_w$ ($N_w < N_W$) being the multispectral band, low-resolution height and low-resolution width, respectively. The point spread function (PSF) ${\bf B} \in \mathbb{R}^{N_HN_W  \times N_HN_W}$ represents the spatial blurring operator, ${\bf D} \in \mathbb{R}^{N_HN_W  \times N_hN_w}$ represents the spatial downsampling operator, ${\bf R} \in \mathbb{R}^{N_b  \times N_B}$ represents the spectral response function (SRF) of the multispectral sensor, and ${\bf E}_y$ and ${\bf E}_z$ are the noises or residuals contained in LR-HSI and HR-MSI. These two degradation models are widely exploited in most of the model-based and deep learning-based HSI super-resolution approaches \cite{yokoya2017hyperspectral,Dian2021RecentAA,Li2022DeepLI}.

Most of the works in HSI super-resolution falls into the model-based or deep learning-based approaches. The model-based approaches mainly raise optimization problems to obtain the desired image, and they have developed many fidelity terms and exploited various prior terms. In the aspect of fidelity terms, many approaches consider using decomposition to depict the intrinsic structure of the target image including unmixing \cite{yokoya2011coupled,lanaras2015hyperspectral}, subspace representation \cite{simoes2015convex,wei2015hyperspectral,liu2020truncated}, dictionary learning \cite{dong2016hyperspectral,veganzones2016hyperspectral,han2020hyperspectral}, and tensor decomposition \cite{li2018fusing,kanatsoulis2018hyperspectral,Xu2020Hyperspectral,Chen2022Hyperspectral}. In the aspect of prior terms, most approaches are devoted to exploiting spatial, spectral or spatial-spectral priors to enforce the desired solution, such as vector total variation \cite{simoes2015convex}, manifold structure \cite{Zhang2018Exploiting}, low-rank representation \cite{Zhang2017Multispectral,Xue2021Spatial}, and tensor regularization \cite{dian2019hyperspectral}.
The deep learning-based approaches are mostly supervised, and they often build an end-to-end deep network to generate the target image by feeding the observed images. Many excellent approaches have been proposed to enhance the fusion ability of the networks, such as residual connection \cite{scarpa2018target}, multiscale branches \cite{yuan2018a}, pyramid structure \cite{zhang2019pan}, attention mechanism \cite{Jiang2020LearningSP}, cross-mode information \cite{Zhang2021SSR}, dense connection \cite{Dong2021Remote}, detail injection \cite{yang2017pannet,xie2019hyperspectral,fu2021deep}, adversarial network \cite{Li2020Hyperspectral,Dong2021GenerativeDN}
and Transformer \cite{Bandara2022HyperTransformerAT,wang2023,xu2023}.

Both the model-based and deep learning-based approaches have their own merits. Some researchers consider combining them to improve the performance of HSI super-resolution. For the model-based approaches, they employ a randomly-initialized neural network as a hand-crafted prior to capture the intrinsic information of HSIs \cite{Zhang2021DeepBH}, or use a trained denoising network as a proximal operator to regulate the results in each iteration \cite{dian2021regularizing}. For the deep learning-based approaches, some build the abstract fusion models and unfold them into several simple operations to guide the construction of networks \cite{MHF,Dong2021Model,wang2023,xu2023,dong2023}, and some use the degradation models to bridge the target image and the observed images and thereby construct unsupervised or zero-shot networks \cite{uSDN,FusionNet,MIAE,Gao2023,guo2023,Dian2023}.

\subsection{Diffusion Models}
\label{sec_dm}
Diffusion models \cite{Luo2022UnderstandingDM,Yang2022DiffusionMA} are the new family of deep generative models that are to learn to model the given data distribution. Different from other generative models, such as generative adversarial networks \cite{GAN} and variational autoencoders \cite{VAE}, diffusion models consist of two iterative processes, forward diffusion process and reverse generative process. For the popular denoising diffusion probabilistic models (DDPMs) \cite{sohl2015deep,DDPM}, given a data distribution ${\bf x}_0 \sim q({\bf x}_0)$, one can define the forward diffusion process $q({\bf x}_t|{\bf x}_{t-1})$ as a Markov chain, which produces latent variables $\{ {\bf x}_t \}_{t=1}^T$ by gradually adding Gaussian noise at every timestep $t$:
\begin{eqnarray}
  q({\bf x}_{1:T}|{\bf x}_0) &=& \prod_{t=1}^T q({\bf x}_t|{\bf x}_{t-1}) \\
  q({\bf x}_t|{\bf x}_{t-1}) &=& \mathcal{N} ({\bf x}_t; \sqrt{1-\beta_t}{\bf x}_{t-1}, \beta_t {\bf I})
\end{eqnarray}
where $\{ \beta_t \}_{t=1}^T$ is a variance schedule with $0 < \beta_1 < \beta_2 < \cdots < \beta_T < 1$. Given a sufficiently large $T$ and well scheduled $\{ \beta_t \}_{t=1}^T$, the forward process produces a nearly isotropic Gaussian distribution ${\bf x}_T$. Under the reparameterization trick, one can sample from $q({\bf x}_t|{\bf x}_0)$ in a closed form:
\begin{equation}
  {\bf x}_t = \sqrt{\bar{\alpha}_t} {\bf x}_0 + \sqrt{1-\bar{\alpha}_t} {\boldsymbol \epsilon}, ~~{\boldsymbol \epsilon} \sim \mathcal{N}({\bf 0}, {\bf I})
\end{equation}
where $\alpha_t=1-\beta_t$ and $\bar{\alpha}_t = \prod_{k=1}^t\alpha_k$. As for the reverse generative process, the estimation of $q({\bf x}_{t-1}|{\bf x}_{t})$ is intractable, and one can turn to use neural networks to estimate it. The reverse generative process is also defined as a Markov chain that consists of a series of parameterized Gaussian transitions $p_{\theta}({\bf x}_{t-1}|{\bf x}_{t})$:
\begin{eqnarray}
  p_{\theta}({\bf x}_{0:T}) &=& p({\bf x}_T) \prod_{t=1}^T p_{\theta}({\bf x}_{t-1}|{\bf x}_{t}) \\
  p_{\theta}({\bf x}_{t-1}|{\bf x}_{t}) &=& \mathcal{N} ({\bf x}_{t-1};
  {\mu}_{\theta} ({\bf x}_t, t), \sigma_t^2 {\bf I}
  )
\end{eqnarray}
where $p({\bf x}_T)=\mathcal{N}({\bf x}_T; {\bf 0}, {\bf I})$, $\mu_{\theta}$ is a learned mean and $\sigma_t$ is a fixed variance. For $\mu_{\theta}({\bf x}_t, t)$, it can be rewritten as:
\begin{equation}
  \mu_{\theta}({\bf x}_t,t) = \frac{1}{\sqrt{\alpha_t}} \left(
  {\bf x}_t - \frac{\beta_t}{\sqrt{1-\bar{\alpha}_t} } {\boldsymbol \epsilon}_{\theta}({\bf x}_t,t)
  \right).
\end{equation}
DDPMs train ${\boldsymbol \epsilon}_{\theta}({\bf x}_t,t)$ by optimizing the following simple denoising-based objective:
\begin{equation}
  \mathcal{L}_{\theta}(\theta) = \mathbb{E}_{{\bf x}_0,{\boldsymbol \epsilon},t}
  \left[
  \| {\boldsymbol \epsilon} - {\boldsymbol \epsilon}_{\theta} (\sqrt{\bar{\alpha}_t} {\bf x}_0 + \sqrt{1-\bar{\alpha}_t} {\boldsymbol \epsilon}, t) \|^2
  \right].
  \label{eq_loss}
\end{equation}
After training, by plugging the learned noise ${\boldsymbol \epsilon}_{\theta}$ into $\mu_{\theta}$, one can simply sample ${\bf x}_{T-1}, {\bf x}_{T-2}, \cdots, {\bf x}_{0}$ by
\begin{equation}
  {\bf x}_{t-1} = \mu_{\theta}({\bf x}_t,t) + \sigma_t {\boldsymbol \epsilon}.
  \label{eq_sample}
\end{equation}

DDPMs and their another related formulation, score-based generative models \cite{song2021score}, have attracted the attention of researchers \cite{Luo2022UnderstandingDM,Yang2022DiffusionMA}. Some approaches focus on strengthening the diffusion models in respect of theoretical foundation \cite{Song2020DenoisingDI,Nichol2021ImprovedDD}, while others employ them to address a variety of challenging real-world applications. In the applications, diffusion models have been successfully applied to many vision tasks, such as image super-resolution \cite{Saharia2023}, object detection \cite{Chen2022DiffusionDetDM}, image restoration \cite{Kawar2022DenoisingDR}, and image fusion \cite{Zhao2023DDFMDD}. Besides these specific applications, there are many strategies proposed to improve the generation quality. Examples are, that using transformer as a backbone for diffusion models outperforms the commonly used U-Net and inherits the excellent scaling properties of the transformer model class \cite{Peebles2022ScalableDM}; or that applying diffusion models in the latent space of pretrained autoencoders allows to reach a near-optimal point between complexity reduction and detail preservation\cite{Rombach2021HighResolutionIS}; or that the reverse generative process in diffusion models can be performed under constraints, making the diffusion models plug-and-play priors \cite{Graikos2022DiffusionMA}; or that applying a learnable encoder to mine the high-level semantics and a diffusion model as the decoder to capture the remaining stochastic variations achieves a meaningful and decodable representation of image data \cite{DiffusionAE}; or that exploiting gradients from a classifier, i.e., classifier guidance, allows to trade off diversity for fidelity, thus improving sample quality in conditional image synthesis \cite{Dhariwal2021DiffusionMB}.

\section{Proposed Approach}
\label{sec_prop}
\subsection{Spectral Diffusion Model}
Deep generative models can learn the complex distributions over high-dimensional data, and thereby randomly generate samples. As one of the most popular models at the moment, diffusion models generate high-quality samples via reverse diffusion processes and have achieved impressive performance in many domains including image \cite{DDPM}, audio \cite{kong2020diffwave}, graph \cite{niu2020permutation} and video \cite{VDM}. Motivated by the success of diffusion models, there is significant interest to investigate the new data modality of HSIs. A straightforward way is to extend the image-based or video-based diffusion models. However, there are two obstacles: insufficient training samples and limited computational resources. In this work, we present solutions on spectrum generation using diffusion models and show that high-quality spectra can be generated. The theoretical foundation of spectral diffusion model has two aspects. One is that like images and videos, a spectrum is a meaningful signal that reflects the absorption properties of different substances. The other is that unlike images and videos, the spectral band number of HSIs is relatively small, reducing the demand of massive training samples and computational resources.

\begin{figure}[!t]
\centering
 {\includegraphics[width=0.5 \textwidth]{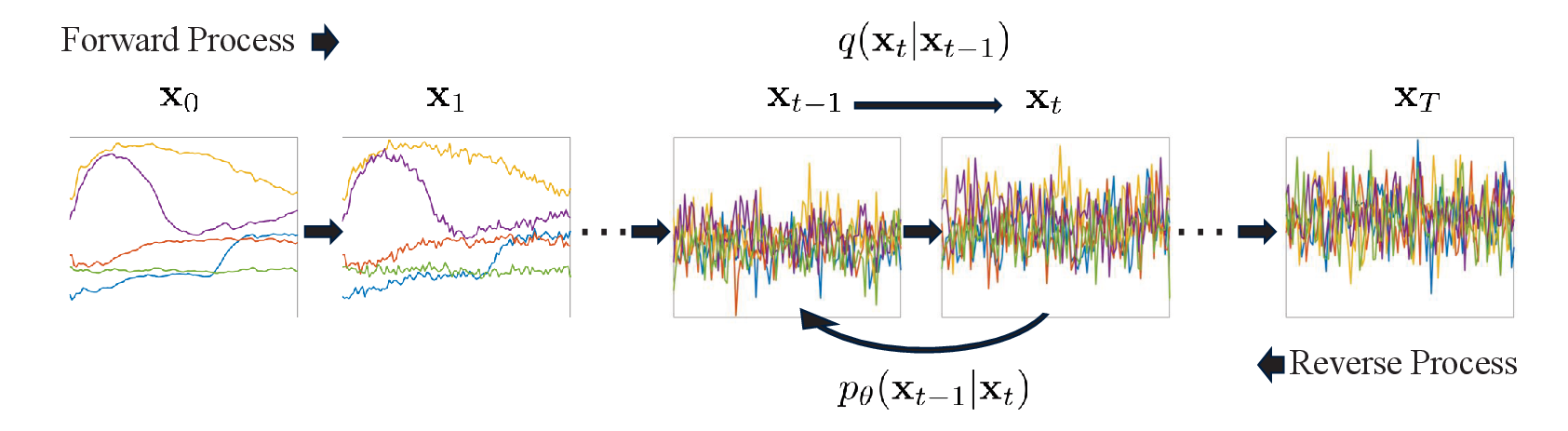}}
\caption{Illustration of spectral diffusion model.}
\label{fig_sdp}
\end{figure}

\begin{figure}[!t]
\centering
 {\includegraphics[width=0.5 \textwidth]{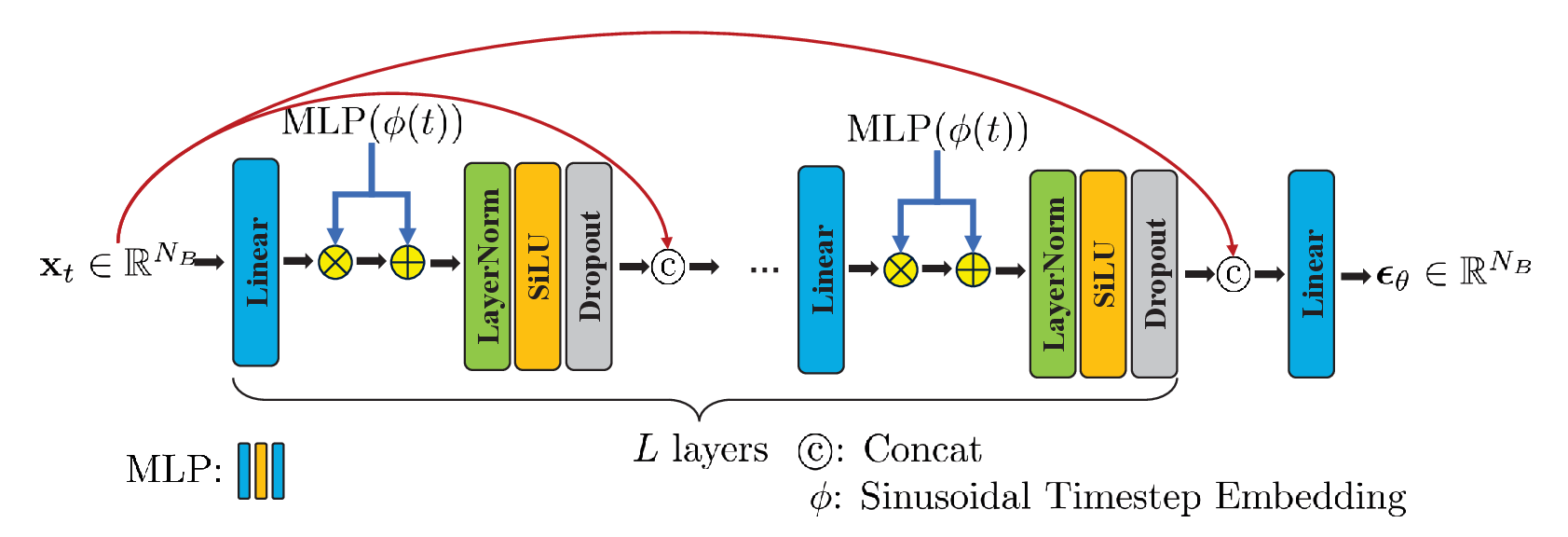}}
\caption{Architecture overview of MLP-based denoising network.}
\label{fig_net}
\end{figure}

\begin{figure}[!t]
\centering
 {\includegraphics[width=0.45 \textwidth]{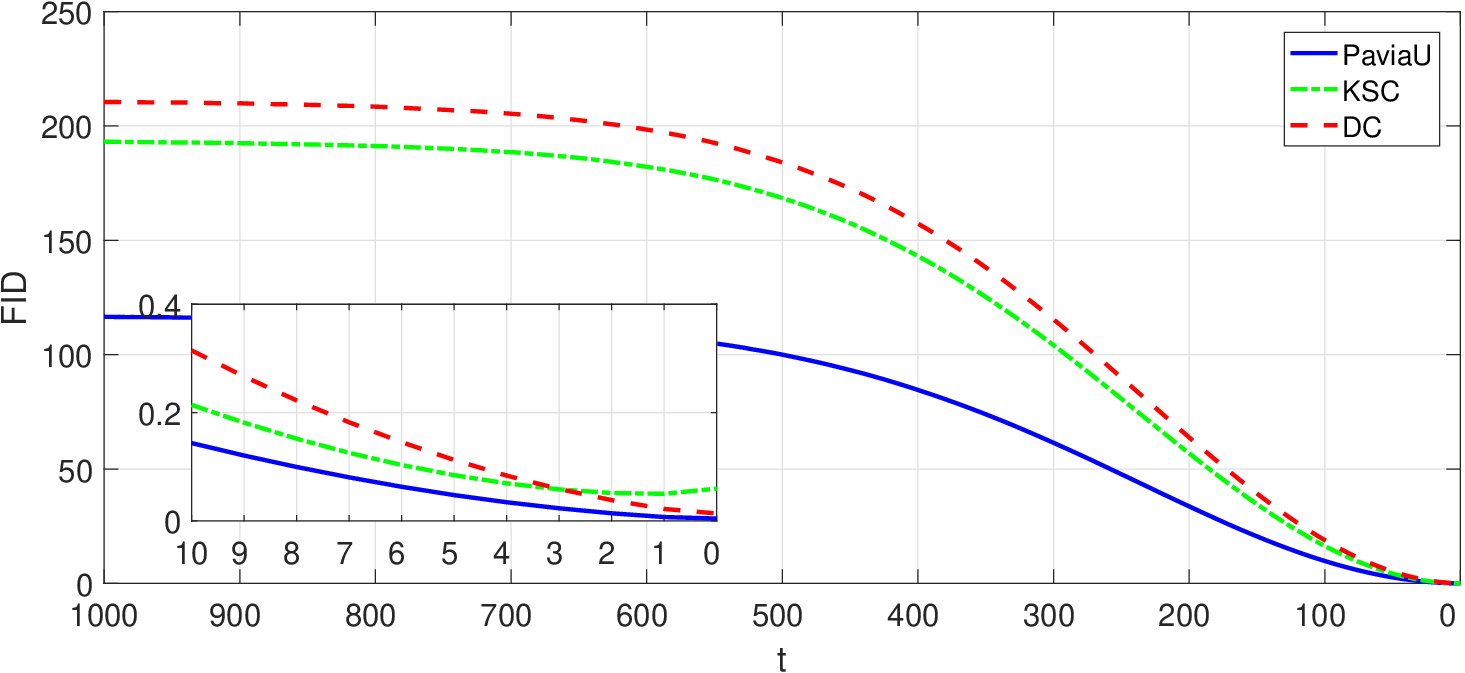}}
\caption{FIDs as a function of timestep $t$ when applied to the given three datasets.}
\label{fig_fid}
\end{figure}

Fig. \ref{fig_sdp} illustrates the proposed spectral diffusion model. In the forward process, it gradually adds Gaussian noise to a spectrum until the spectrum becomes random noise. In the reverse process, it starts from Gaussian noise and learns to denoise a spectrum by modeling the spectral data distribution. The spectral diffusion model is designed by following the standard diffusion model formalism (i.e., DDPMs) described in Section \ref{sec_dm}. The difference is that it uses a MLP-based denoising network architecture. In previous works on diffusion models, the standard architecture of denoising networks is a U-Net \cite{UNet} with time embedding and spatial attentions. One can adjust the U-Net to match spectral data, but it is suboptimal due to the relatively low spectral dimension and the downstream computational requirements. Fig. \ref{fig_net} shows the MLP-based denoising network architecture. The adopted network consists of $L$ layers. Each layer outputs a $N_{out}$-dimensional feature vector and then concatenates it with the input via a skip connection. Similar to other diffusion models, this denoising network is conditioned on the timestep $t$, which performs sinusoidal timestep embedding followed by a MLP layer. After training the denoising network under the loss (\ref{eq_loss}), one can generate spectra gradually via (\ref{eq_sample}).

In image diffusion models, Frechet inception distance (FID) \cite{FID} is most often used to evaluate the quality of samples. We have found that FID is also suitable for spectral data. In the reverse generative process, the quality of generation spectra gradually becomes better, as the denoising steps are performed. Fig. \ref{fig_fid} reports the values of FID at different timestep $t$ during the generative process.

\subsection{HSI Super-Resolution with Spectral Diffusion Prior}
The goal of HSI super-resolution is to generate the target HR-HSI ${\bf X}$ by fusing the LR-HSI ${\bf Y}$ and HR-MSI ${\bf Z}$. In the framework of maximum a posteriori, the problem of HSI super-resolution can be written as
\begin{equation}
  \max_{\bf X} p({\bf X} | {\bf Y}, {\bf Z}).
  \label{eq_map}
\end{equation}
Using the Bayes formula, the posterior $p({\bf X} | {\bf Y}, {\bf Z})$ can be rewritten as:
\begin{equation}
  p({\bf X} | {\bf Y}, {\bf Z}) = \frac{p({\bf Y}, {\bf Z} | {\bf X})p({\bf X})} {p({\bf Y}, {\bf Z})}.
\end{equation}
Then, the problem (\ref{eq_map}) is equivalent as:
\begin{equation}
  \min_{\bf X} -\log p({\bf Y}, {\bf Z} | {\bf X}) - \log p({\bf X}),
  \label{eq_opt}
\end{equation}
where the first term is the log-likelihood that depicts the degradation processes of HSI super-resolution, and the irrelevant term $p({\bf Y}, {\bf Z})$ is dropped. In the optimization, one can write $-\log p({\bf Y}, {\bf Z} | {\bf X})$ as the following formulation according to (\ref{eq_hsi}) and (\ref{eq_msi}),
\begin{equation}
  \mathcal{L}_X({\bf X}) = \lambda \|{\bf Y} - {\bf X}{\bf BD}\|_F^2 + \|{\bf Z} - {\bf R}{\bf X}\|_F^2,
\end{equation}
where $\lambda > 0$ is a balance parameter, and $\| \cdot \|_F$ represents the Frobenius norm. In (\ref{eq_opt}), $- \log p({\bf X})$ is the prior (termed as regularization term) that contributes to recovering the target ${\bf X}$.

Various regularization terms have been proposed for $- \log p({\bf X})$ in previous works \cite{yokoya2011coupled,lanaras2015hyperspectral,simoes2015convex,wei2015hyperspectral,liu2020truncated,dong2016hyperspectral,veganzones2016hyperspectral,han2020hyperspectral,li2018fusing,kanatsoulis2018hyperspectral,Xu2020Hyperspectral,Chen2022Hyperspectral,Zhang2018Exploiting,Zhang2017Multispectral,Xue2021Spatial,dian2019hyperspectral}. Different from them, we assume the set of spectral vectors ${\bf x} \in \mathbb{R}^{N_B}$ in ${\bf X}$ is independent identically distributed, i.e., $- \log p({\bf X}) = -\sum_{{\bf x} \in {\bf X}} \log p({\bf x})$, and then exploit a spectral diffusion prior for HSI super-resolution. To use the spectral diffusion model as a prior over ${\bf x}$, we assume the spectrum ${\bf x} = {\bf x}_0$ follows a given spectral data distribution $q({\bf x}_0)$, and consider the joint distribution of all its states in the Markov chain, i.e.,
\begin{eqnarray}
  \log q({\bf x}_{0:T} | {\bf x})
  &=& \log \left( q({\bf x}_T | {\bf x}) \prod_{t=1}^T q( {\bf x}_{t-1} | {\bf x}_{t}, {\bf x}) \right) \\
  &=& \log q({\bf x}_T | {\bf x}) + \sum_{t=1}^T \log q( {\bf x}_{t-1} | {\bf x}_{t}, {\bf x}) ~~~~~\label{eq_state}
\end{eqnarray}
The goal of designing the spectral diffusion prior is to embed the knowledge of $\log q({\bf x}_{0:T} | {\bf x})$ into the optimization process. In (\ref{eq_state}), the first term $\log q({\bf x}_T | {\bf x})$ can be omitted as its information is unavailable. The second term $\log q( {\bf x}_{t-1} | {\bf x}_{t}, {\bf x})$ is the true logarithmic posteriori distribution that depicts the transition of two neighboring Markov states given ${\bf x}$. In the spectral diffusion model, the knowledge is stored in the neural network ${\boldsymbol \epsilon}_{\theta}({\bf x}_t,t)$ or rather in $p_{\theta}({\bf x}_{t-1} | {\bf x}_{t})$, and the true posteriori distribution $q( {\bf x}_{t-1} | {\bf x}_{t}, {\bf x}_0)$ is approximated by $p_{\theta}({\bf x}_{t-1} | {\bf x}_{t})$. Thus, we introduce a constraint term to embed the knowledge of the spectral diffusion model, and the optimization problem (\ref{eq_opt}) can be written as
\begin{eqnarray}
  \min_{\bf X} && \mathcal{L}_X({\bf X}) = \lambda \|{\bf Y} - {\bf X}{\bf BD}\|_F^2 + \|{\bf Z} - {\bf R}{\bf X}\|_F^2 \nonumber \\
  {\rm s.t.} && q( {\bf x}_{t-1} | {\bf x}_{t}, {\bf x}) = p_{\theta} ( {\bf x}_{t-1} | {\bf x}_{t}) \label{eq_opt_1}\\
  && t\in\{1, \cdots, T\}, ~~ {\bf x} \in {\bf X} \nonumber
\end{eqnarray}
The optimization problem (\ref{eq_opt_1}) is a constrained one, and we can rewrite it as the following unconstrained form
\begin{equation}
  \min_{\bf X} \mathcal{L}_X({\bf X}) + \gamma \sum_{t,{\bf x}} \mathcal{L}_{kl} (q( {\bf x}_{t-1} | {\bf x}_{t}, {\bf x}) || p_{\theta} ( {\bf x}_{t-1} | {\bf x}_{t})), \label{eq_opt_2}
\end{equation}
where $\gamma >0$ is a regularization parameter, and $\mathcal{L}_{kl}$ denotes the Kullback-Leibler (KL) divergence. Under the assumption of ${\bf x}$, the second term of (\ref{eq_opt_2}) can be rewritten as the form of (\ref{eq_loss}) using the reparameterization trick,
\begin{equation}
  \mathcal{L}_{\theta}({\bf x}, t) =  \mathbb{E}_{{\boldsymbol \epsilon}}
  \left[
  \| {\boldsymbol \epsilon} - {\boldsymbol \epsilon}_{\theta} (\sqrt{\bar{\alpha}_t} {\bf x} + \sqrt{1-\bar{\alpha}_t} {\boldsymbol \epsilon}, t) \|^2
  \right],
  \label{eq_error}
\end{equation}
where the weight that is a function of the noise schedule is omitted. Then the problem (\ref{eq_opt_2}) can be formulated as
\begin{equation}
  \min_{\bf X} \mathcal{L}_X({\bf X}) + \gamma \sum_{t,{\bf x}} \mathcal{L}_{\theta}({\bf x}, t).
  \label{eq_final}
\end{equation}

\subsection{Optimization}
The final objective (\ref{eq_final}) consists of an ordinary fidelity term and a spectral diffusion regularization term. The regularization term is originally used to train the network ${\boldsymbol \epsilon}_{\theta}$ of the spectral diffusion model, and its network parameters $\theta$ are fixed in solving (\ref{eq_final}). To solve (\ref{eq_final}), we should optimize the objective with respect to ${\bf X}$ through sampling ${\boldsymbol \epsilon} \sim \mathcal{N}({\bf 0}, {\bf 1})$ in the summands over $t$.

The spectral diffusion model gradually generates spectra using (\ref{eq_sample}) during the sampling process. Intuitively, a large $t$ should coarsely generate ${\bf x}_0=({\bf x}_t - \sqrt{1-{\bar \alpha}_t} {\boldsymbol \epsilon}_{\theta}) / \sqrt{{\bar \alpha}_t}$, and it is impractical to optimize (\ref{eq_final}) simultaneously for all $t$. Similar to the sampling process, we sequentially solve the $t$th subproblem of (\ref{eq_final}) from $T$ to $1$, so as to match the iterative solutions in optimization. Specifically, we treat ${\bf X}$ as trainable parameters, adopt the Adam as the optimizer and perform the gradient update $K$ times for each $t$-subproblem. The complete process of the proposed SDP is summarized in Algorithm \ref{alg_sdp}.

\begin{algorithm}
\caption{SDP}
\begin{algorithmic}[1]
\State \textbf{Input}:
LR-HSI ${\bf Y}$, HR-MSI ${\bf Z}$, clean hyperspectral pixels ${\tilde{\bf X}}$, PSF ${\bf B}$, SRF ${\bf R}$, regularization parameters $\lambda$ and $\gamma$, learning rate $\mu$.
\State Train the denoising network ${\boldsymbol \epsilon}_{\theta}$ of spectral diffusion model by feeding ${\tilde{\bf X}}$.
\State Initialize the parameters ${\bf X}$.
\For{$t = T : 1$}
\For{$k = 1 : K$}
\State Sample ${\boldsymbol \epsilon} \in \mathbb{R}^{N_B \times N_HN_W} \sim \mathcal{N}({\bf 0}, {\bf 1})$
\State ${\bf X}_t = \sqrt{\bar{\alpha}_t} {\bf X} + \sqrt{1-\bar{\alpha}_t} {\boldsymbol \epsilon}$
\State Compute ${\boldsymbol \epsilon}_{\theta} \in \mathbb{R}^{N_B \times N_HN_W}$ by feeding ${\bf X}_t$ and $t$
\State ${\bf X} \leftarrow {\bf X} - \mu \nabla_{\bf X} \left[ \mathcal{L}_X({\bf X}) + \gamma \sum_{{\bf x}} \mathcal{L}_{\theta}({\bf x}, t) \right]$
\EndFor
\EndFor
\State \textbf{Output}: HR-HSI ${\bf X}$.
\end{algorithmic}
\label{alg_sdp}
\end{algorithm}

\section{Experimental Results and Analysis}
\label{sec_ex}
This section conducts experiments on both synthetic and real datasets to evaluate the performance of the proposed SDP. All datasets are scaled to the range $[0, 1]$. For the synthetic datasets, four metrics are used to assess the quality of the fused images, namely, peak signal-noise-ratio (PSNR), spectral angle mapper (SAM), root
mean squared error (RMSE), relative dimensionless global error in synthesis (ERGAS), and universal image quality index (UIQI) \cite{loncan2015hyperspectral,yokoya2017hyperspectral,Dian2021RecentAA}. For the real dataset, three metrics without reference are adopted to assess the quality of the fusion results, namely, spectral distortion index $D_{\lambda}$, spatial distortion index $D_{s}$ and quality with no reference (QNR) \cite{QNR,Vivone2021A}.

\subsection{Synthetic Datasets}
Given three real-life HSIs, University of Paiva (PaviaU), Kennedy Space Center (KSC), and Washington DC Mall (DC), we take the source image as the reference image and generate the corresponding two observation images LR-HSI and HR-MSI, according to the Wald's protocol \cite{ranchin2000fusion}. The three real-life HSIs are listed as follows.
\begin{enumerate}
\item The PaviaU dataset is captured by the Reflective Optics System Imaging Spectrometer (ROSIS). This image has 115 spectral bands covering the spectral range 0.43--0.86 $\mu m$, and has $610 \times 340$ pixels with a spatial resolution of 1.3 $m$ per pixel. After removal of noisy bands, it has 103 spectral bands remained. We crop the up-left $512 \times 256$-pixel part for experiments.
\item The KSC dataset is captured by the Airborne Visible/Infrared Imaging Spectrometer (AVIRIS). This image has 224 spectral bands covering the spectral range 0.4--2.5 $\mu m$, and has $512 \times 614$ pixels with a spatial resolution of 18 $m$ per pixel. The number of its spectral bands is reduced to 176 by removing water absorption bands. We crop the up-left $512 \times 256$-pixel part for experiments.
\item The DC dataset is captured by the Hyperspectral digital imagery collection experiment (HYDICE) image. This image has 210 spectral bands covering the spectral range 0.4--2.4 $\mu m$, and has $1208 \times 307$ pixels with a spatial resolution of about 2.8 $m$ per pixel. There are 191 bands left by removing bands in the specific region, where the atmosphere is opaque. We crop the up-left $512 \times 256$-pixel part for experiments.
\end{enumerate}

We divide the cropped image into two $256 \times 256$-pixel parts. The top one is used as the clean hyperspectral pixels $\tilde{\bf X}$, and the bottom one is used as the reference image ${\bf X}$. The synthetic processes of the observation images LR-HSI and HR-MSI can be depicted by (\ref{eq_hsi}) and (\ref{eq_msi}), respectively. For the LR-HSI, the PSF ${\bf B}$ is achieved by spatially blurring each band of the reference image, where a Gaussian blur of size $15 \times 15$ with mean 0 and standard deviation 3.40 is adopted. The operator ${\bf D}$ is achieved by spatially downsampling the blurred image with a factor of 8 in two directions. The noise ${\bf E}_y$ is achieved by adding 20 dB Gaussian noise to the LR-HSI. For the HR-MSI, the SRF ${\bf R} \in \mathbb{R}^{4\times N_B}$ is derived from the spectral response of the IKONOS satellite,  according to the spectral response profiles of the RGB and NIR bands. A Gaussian noise with 30 dB is added to the HR-MSI.

\subsection{Implementation Details and Influence of Parameters}
\begin{table}[!t]
\caption{Implementation details of spectral diffusion model}
\label{tab_net}
\centering
\tabcolsep = 4.0pt
\begin{tabular}
{l|c}
\hline\hline
Parameter                 & Value \\
\hline\hline
MLP layers $L$            & 4 \\
MLP hidden size $N_{out}$ & 512 \\
Time embedding dim.       & 64 \\
Dropout rate              & 0.001 \\
Time steps $T$            & 1000 \\
$\beta$ scheduler        & Linear\\
$\beta_1$, $\beta_T$        & 0.0001, 0.02\\
Optimizer                 & Adam \\
Batch size                & 512 \\
Epoches                   & 30000 \\
Learning rate             & 0.01 \\
Learning rate scheduler   & $0.001 \times \max(1000 - epoch / 10, 1)$ \\
\hline\hline
\end{tabular}
\end{table}

\begin{figure*}[!t]
\centering
\subfigure[] {\includegraphics[width=0.3 \textwidth]{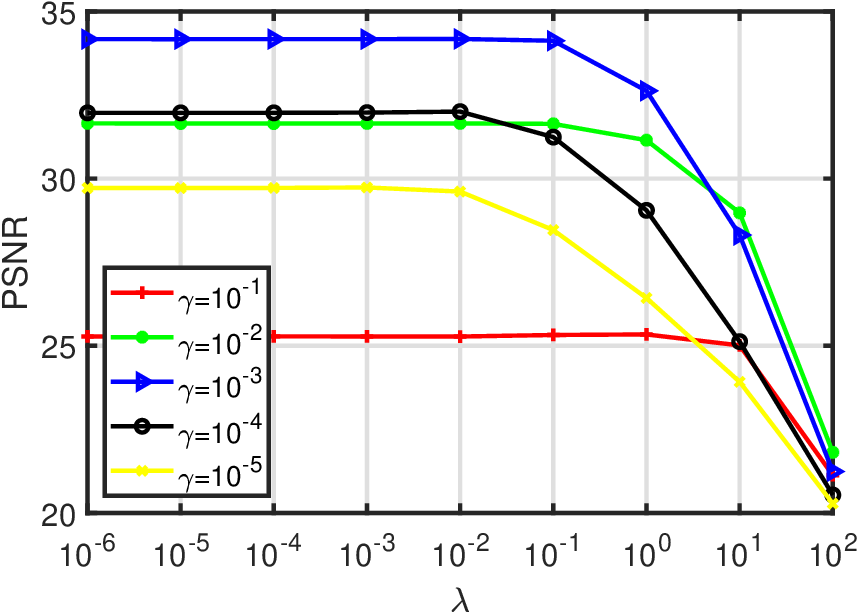}}~~~
\subfigure[] {\includegraphics[width=0.3 \textwidth]{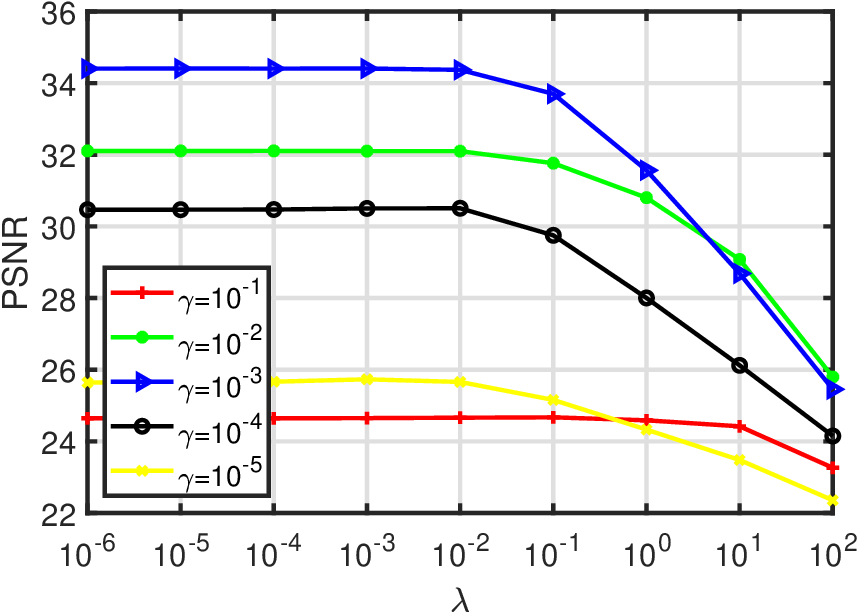}}~~~
\subfigure[] {\includegraphics[width=0.3 \textwidth]{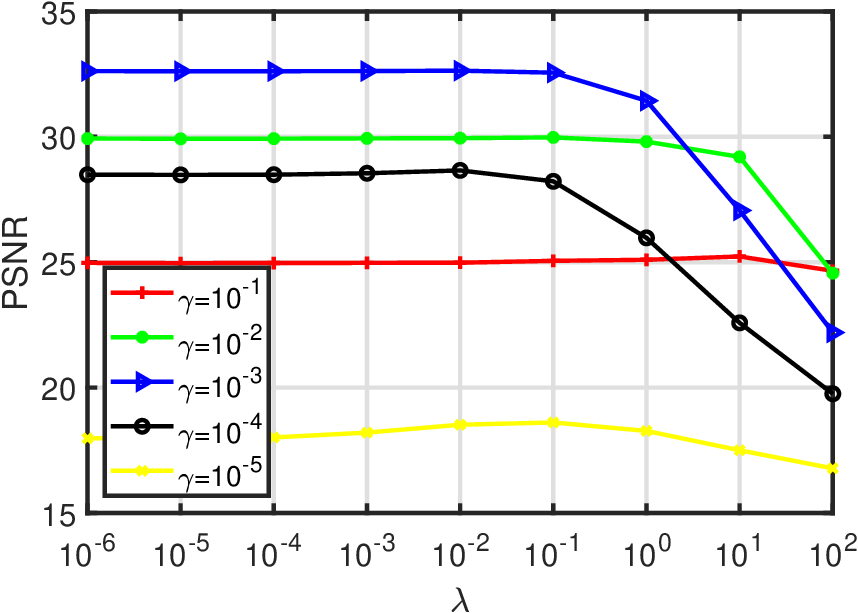}}
\caption{PSNR with respect to parameters $\lambda$ and $\gamma$. (a) PaviaU dataset. (b) KSC dataset. (c) DC dataset.}
\label{fig_param}
\end{figure*}

We implement the spectral diffusion model and SDP in the PyTorch framework. The implementation details of the proposed spectral diffusion model are provided in Table \ref{tab_net}. For SDP, the inner loop $K$ is set to be 3, and the learning rate $\mu$ is set as 0.001, 0.001, 0.0025 for the PaviaU, KSC and DC datasets, respectively.

There are two parameters $\lambda$ and $\gamma$ that need to be tuned in the proposed SDP. This set of experiments is to investigate them and show their impact on the fusion results. Fig. \ref{fig_param} illustrates the influence of these two parameters in terms of PSNR. It can be observed that the results of $\gamma=10^{-3}$ are the best for all datasets. The choice of $\lambda$ has a wide optimal range in $[10^{-6}, 10^{-1}]$, which indirectly demonstrates the effectiveness of the proposed spectral diffusion prior. Thus, $\lambda$ and $\gamma$ are eventually set as $10^{-1}$ and $10^{-3}$ for all datasets.

\subsection{Experiment Results on Synthetic Datasets}
\label{sec_syn}
\subsubsection{Relationship between FID and Fusion Accuracy}
\begin{figure*}[!t]
\centering
\subfigure[] {\includegraphics[width=0.3 \textwidth]{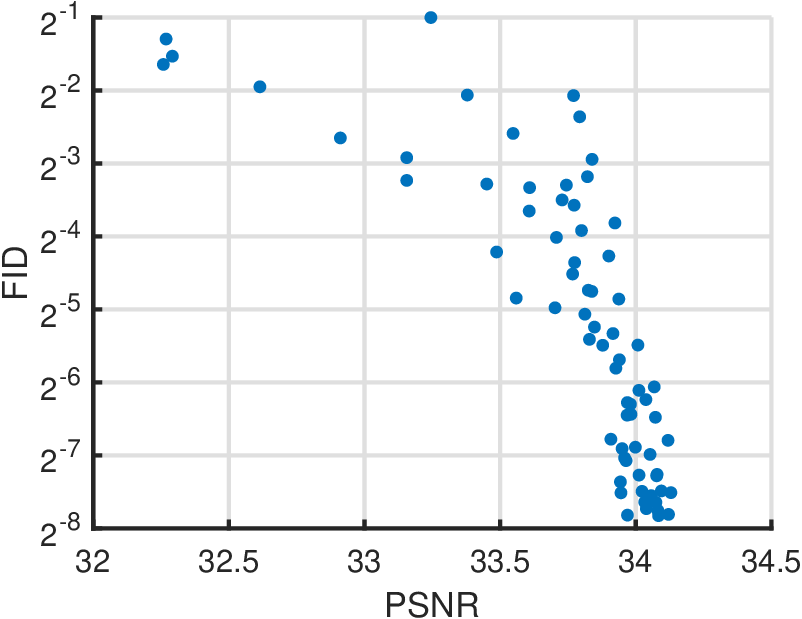}}~~~
\subfigure[] {\includegraphics[width=0.3 \textwidth]{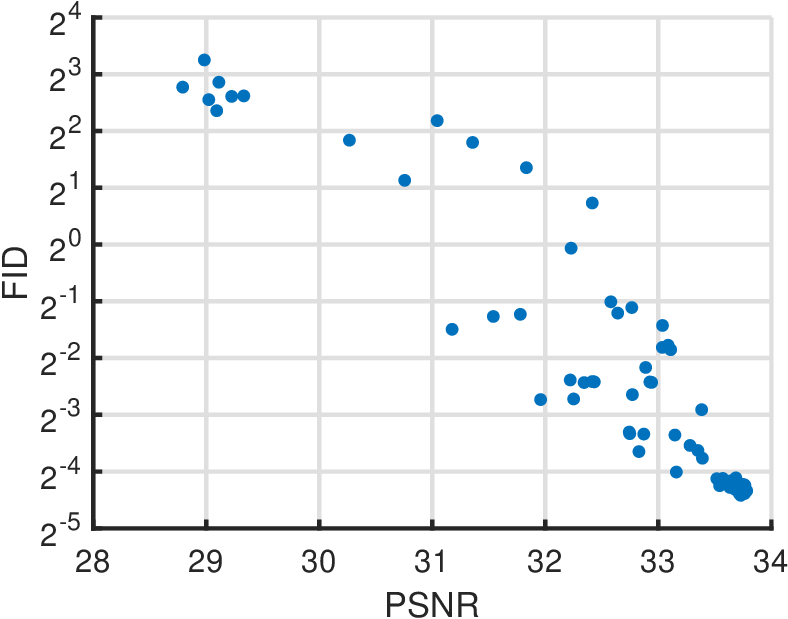}}~~~
\subfigure[] {\includegraphics[width=0.3 \textwidth]{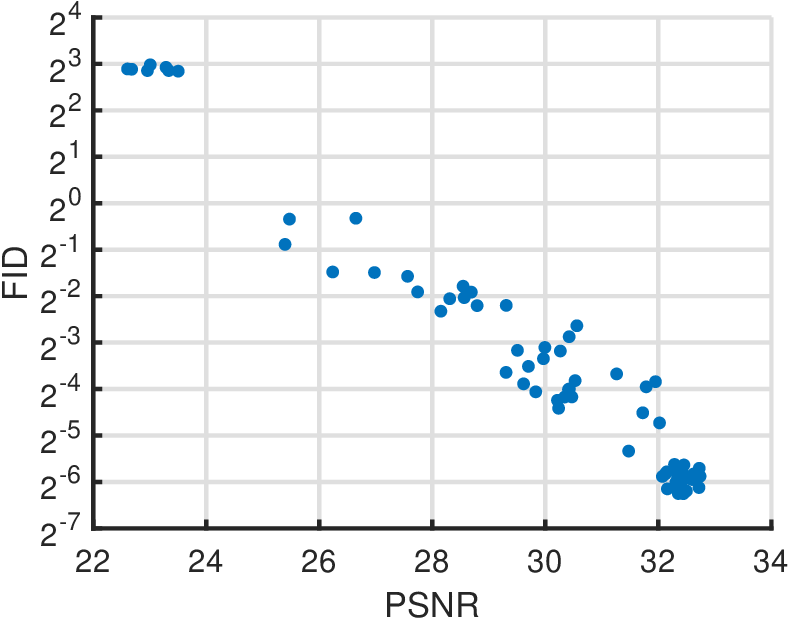}}
\caption{Illustration the relationship between FID and PSNR. (a) PaviaU dataset. (b) KSC dataset. (c) DC dataset.}
\label{fig_fid_psnr}
\end{figure*}
This set of experiments is to show how the spectrum generation quality FID affects the fusion accuracy. By varying MLP layers $L$, hidden size $N_{out}$, and training epoch, we train 71 different spectral diffusion models and thereby compute the FIDs of their generated spectra. Then, for each model, we perform SDP by adopting it as a prior. Fig. \ref{fig_fid_psnr} illustrates the FIDs of these spectral diffusion models and the PSNRs of their corresponding fusion results. It is clear that in the general trend, the higher the generation quality of the spectral diffusion model, the better the fusion result with it as a prior.

\subsubsection{Comparison With the State of the Art}
\begin{table}[!t]
\caption{Comparison of different methods on the PaviaU dataset (the best values are marked in bold)}
\label{tab_pavia}
\centering
\begin{tabular}
{l|c|c|c|c|c}
\hline\hline
Method & PSNR & SAM & RMSE & ERGAS & UIQI \\
\hline
Best Values & $+\infty$ & 0 & 0 & 0 & 1 \\
\hline\hline
  Baseline &27.88 &7.98 &0.061 &1.921 &0.934 \\
       CSU &29.98 &4.34 &0.033 &1.217 &0.971 \\
    HySure &32.05 &3.81 &0.027 &1.029 &0.979 \\
      NSSR &29.74 &4.57 &0.035 &1.206 &0.971 \\
      HCTR &32.05 &4.06 &0.029 &1.045 &0.979 \\
    CNNFUS &32.37 &3.54 &0.026 &0.981 &0.980 \\
      MIAE &32.27 &3.93 &0.027 &1.001 &0.978 \\
       SDP &\textbf{34.12} &\textbf{3.07} &\textbf{0.022} &\textbf{0.825} &\textbf{0.984} \\
\hline\hline
\end{tabular}
\end{table}

\begin{figure*}[!t]
\centering
{\includegraphics[width=\myimgsize in]{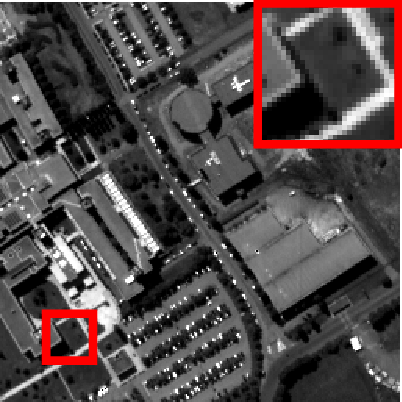}}
{\includegraphics[width=\myimgsize in]{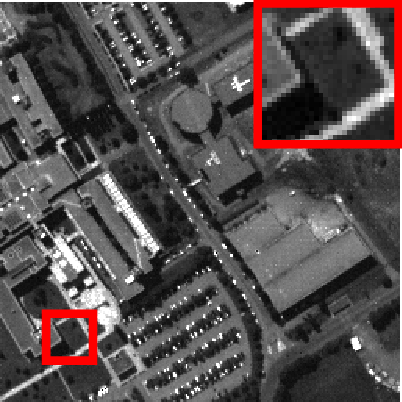}}
{\includegraphics[width=\myimgsize in]{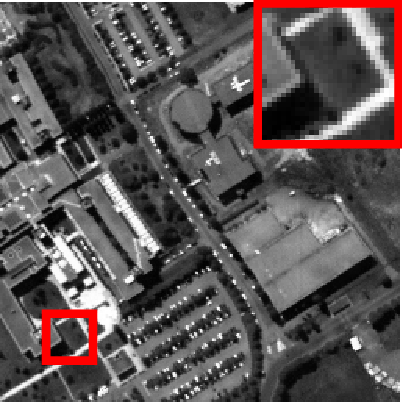}}
{\includegraphics[width=\myimgsize in]{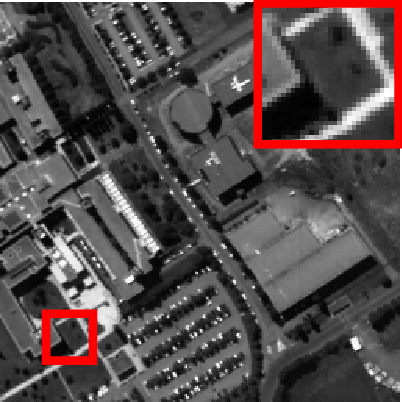}}
{\includegraphics[width=\myimgsize in]{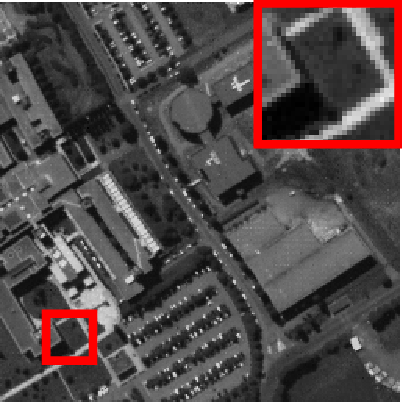}}
{\includegraphics[width=\myimgsize in]{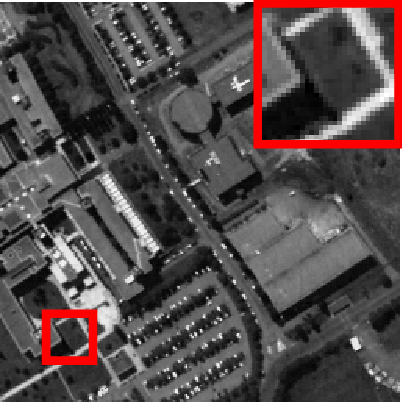}}
{\includegraphics[width=\myimgsize in]{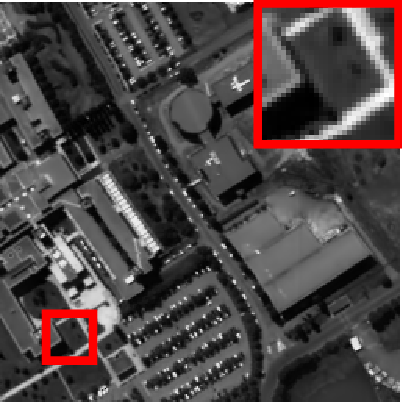}}
{\includegraphics[width=\myimgsize in]{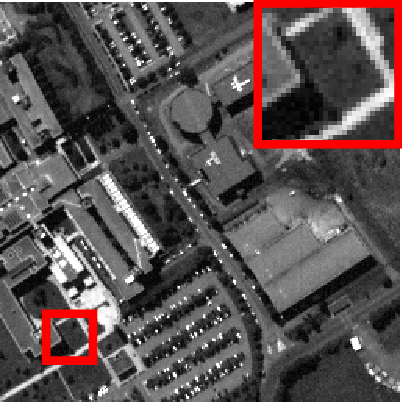}}
{\includegraphics[width=\myimgsize in]{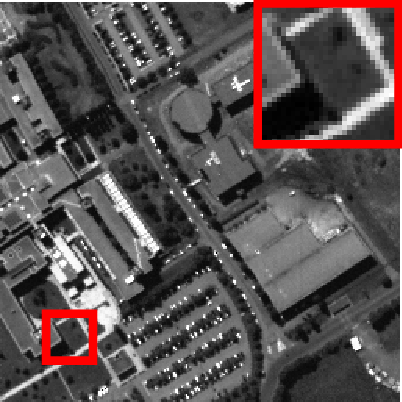}} \\
\subfigure[] {\includegraphics[width=\myimgsize in]{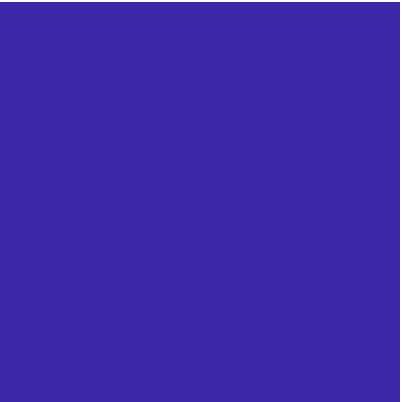}}
\subfigure[] {\includegraphics[width=\myimgsize in]{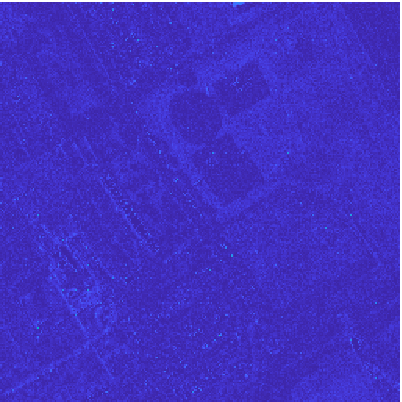}}
\subfigure[] {\includegraphics[width=\myimgsize in]{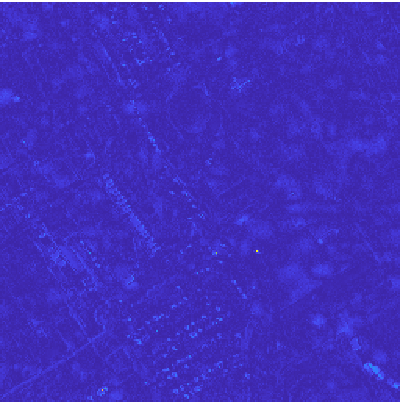}}
\subfigure[] {\includegraphics[width=\myimgsize in]{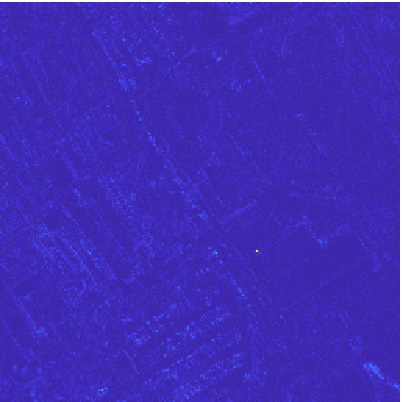}}
\subfigure[] {\includegraphics[width=\myimgsize in]{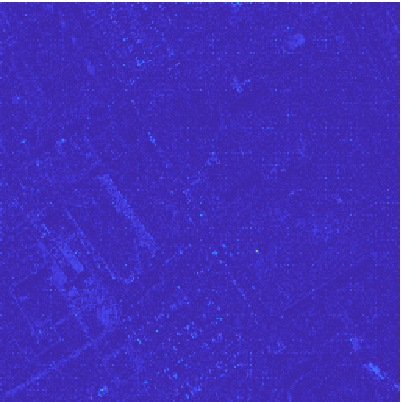}}
\subfigure[] {\includegraphics[width=\myimgsize in]{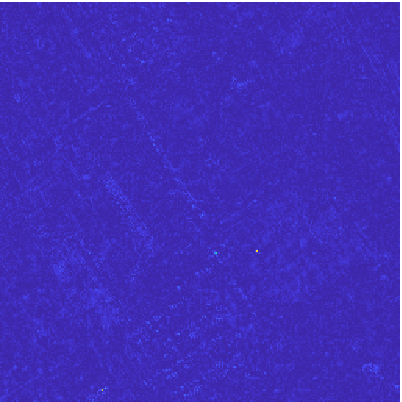}}
\subfigure[] {\includegraphics[width=\myimgsize in]{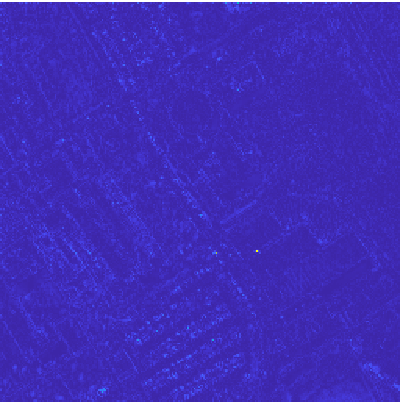}}
\subfigure[] {\includegraphics[width=\myimgsize in]{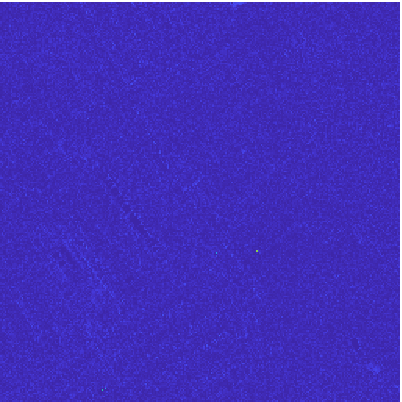}}
\subfigure[] {\includegraphics[width=\myimgsize in]{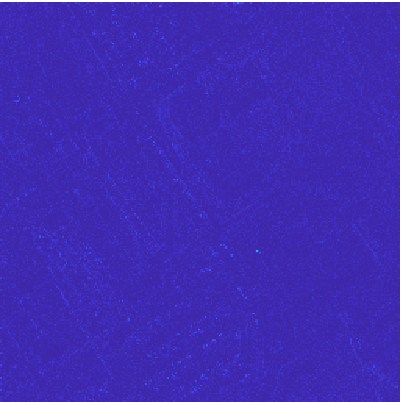}}
\caption{Fused images and error maps at the 30th band when applied to the PaviaU dataset.
(a) Reference image. (b) Baseline. (c) CSU. (d) HySure.
(e) NSSR. (f) HCTR. (g) CNNFUS.
(h) MIAE. (i) SDP.}
\label{fig_pavia}
\end{figure*}

\begin{table}[!t]
\caption{Comparison of different methods on the KSC dataset (the best values are marked in bold)}
\label{tab_ksc}
\centering
\begin{tabular}
{l|c|c|c|c|c}
\hline\hline
Method & PSNR & SAM & RMSE & ERGAS & UIQI \\
\hline
Best Values & $+\infty$ & 0 & 0 & 0 & 1 \\
\hline\hline
  Baseline &21.97 &19.34 &0.150 &8.284 &0.512 \\
       CSU &27.02 &6.76 &0.047 &2.614 &0.695 \\
    HySure &31.98 &4.56 &0.029 &1.929 &\textbf{0.826} \\
      NSSR &28.70 &7.22 &0.040 &2.527 &0.707 \\
      HCTR &31.56 &6.39 &0.032 &2.211 &0.755 \\
    CNNFUS &30.93 &4.38 &0.032 &1.953 &0.821 \\
      MIAE &30.27 &6.51 &0.038 &2.375 &0.726 \\
       SDP &\textbf{33.70} &\textbf{4.11} &\textbf{0.026} &\textbf{1.787} &0.805 \\
\hline\hline
\end{tabular}
\end{table}

\begin{figure*}[!t]
\centering
{\includegraphics[width=\myimgsize in]{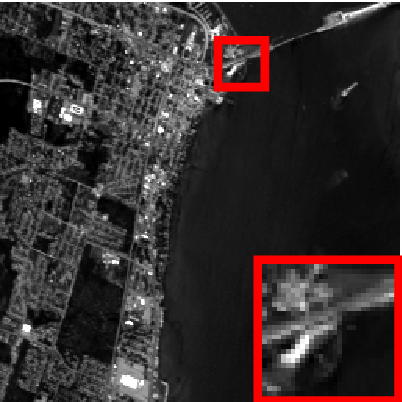}}
{\includegraphics[width=\myimgsize in]{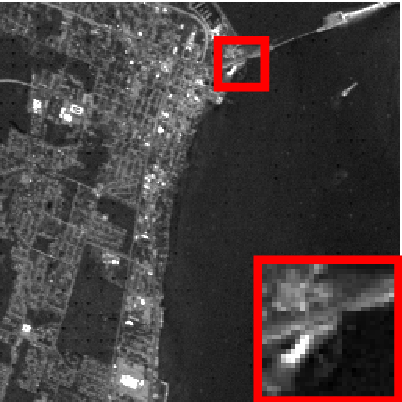}}
{\includegraphics[width=\myimgsize in]{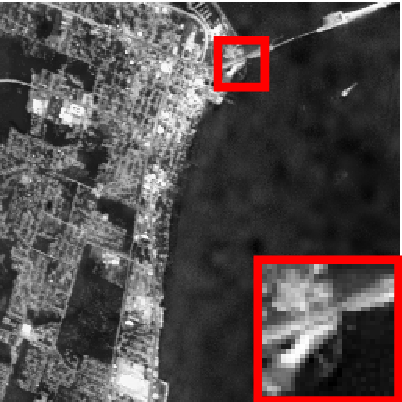}}
{\includegraphics[width=\myimgsize in]{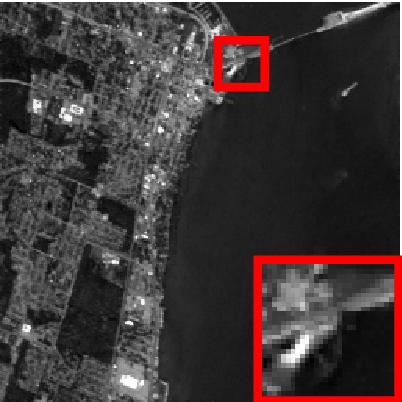}}
{\includegraphics[width=\myimgsize in]{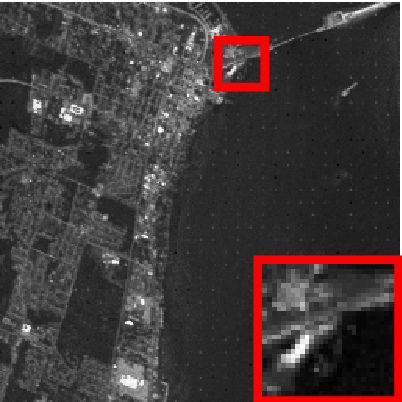}}
{\includegraphics[width=\myimgsize in]{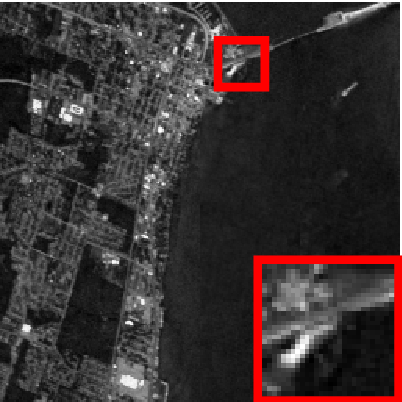}}
{\includegraphics[width=\myimgsize in]{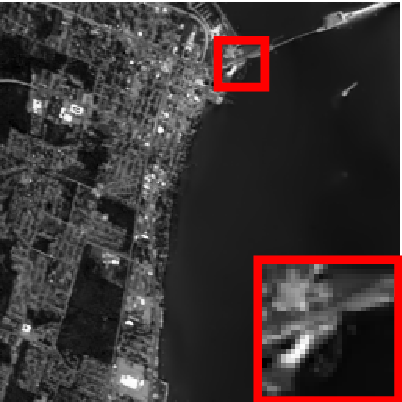}}
{\includegraphics[width=\myimgsize in]{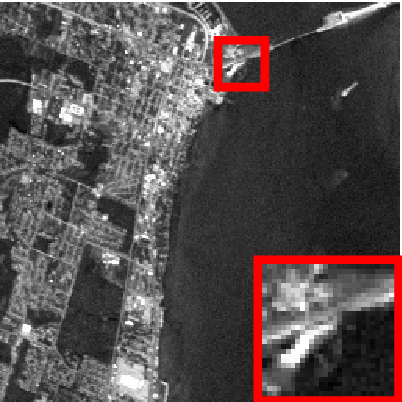}}
{\includegraphics[width=\myimgsize in]{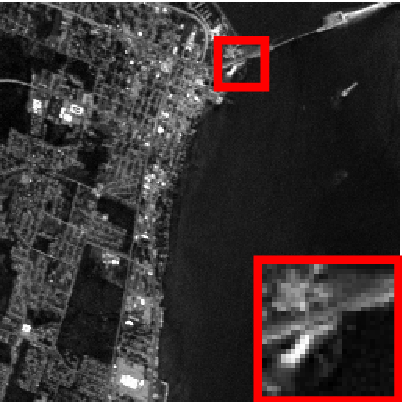}} \\
\subfigure[] {\includegraphics[width=\myimgsize in]{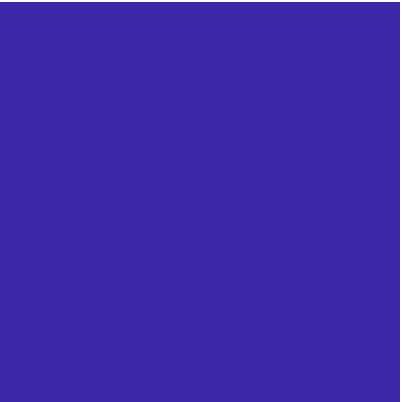}}
\subfigure[] {\includegraphics[width=\myimgsize in]{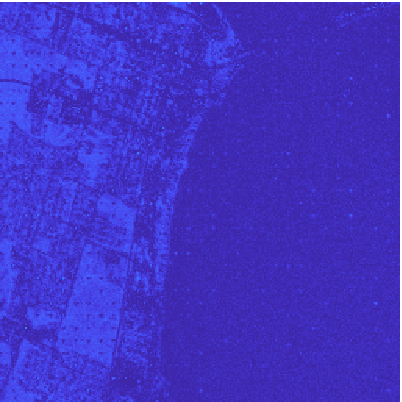}}
\subfigure[] {\includegraphics[width=\myimgsize in]{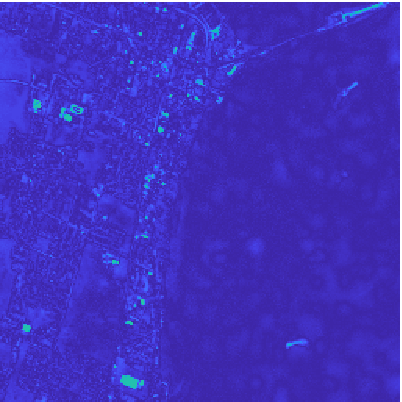}}
\subfigure[] {\includegraphics[width=\myimgsize in]{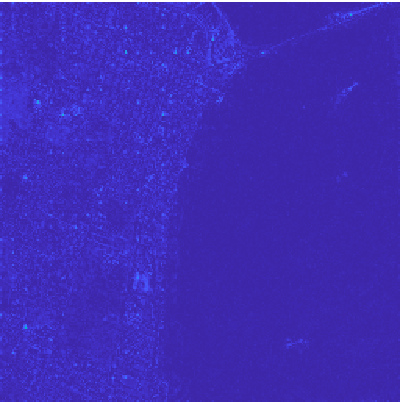}}
\subfigure[] {\includegraphics[width=\myimgsize in]{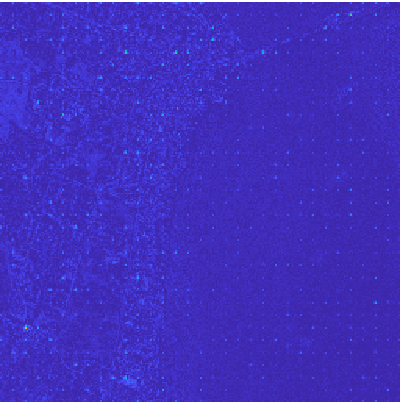}}
\subfigure[] {\includegraphics[width=\myimgsize in]{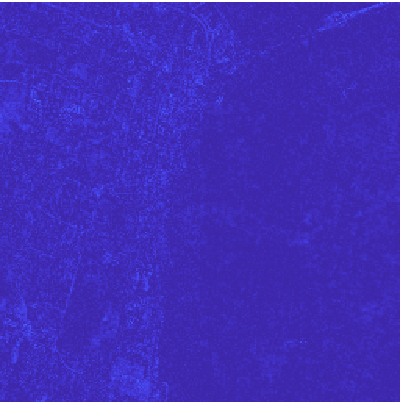}}
\subfigure[] {\includegraphics[width=\myimgsize in]{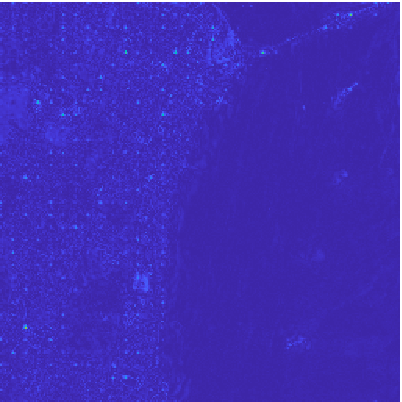}}
\subfigure[] {\includegraphics[width=\myimgsize in]{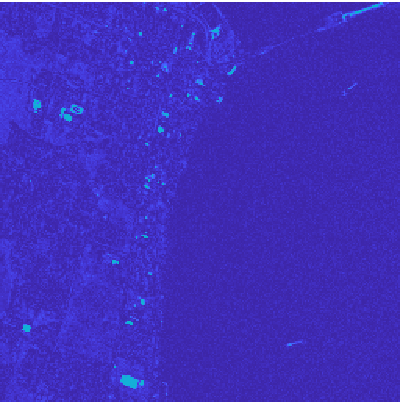}}
\subfigure[] {\includegraphics[width=\myimgsize in]{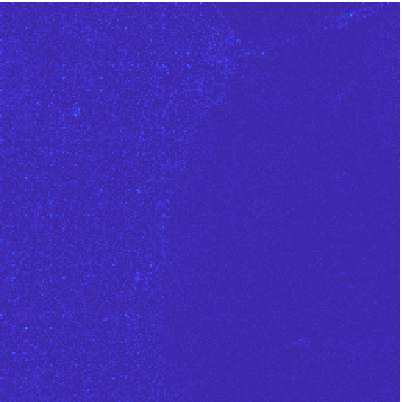}}
\caption{Fused images and error maps at the 30th band when applied to the KSC dataset.
(a) Reference image. (b) Baseline. (c) CSU. (d) HySure.
(e) NSSR. (f) HCTR. (g) CNNFUS.
(h) MIAE. (i) SDP.}
\label{fig_ksc}
\end{figure*}

\begin{table}[!t]
\caption{Comparison of different methods on the DC dataset (the best values are marked in bold)}
\label{tab_dc}
\centering
\begin{tabular}
{l|c|c|c|c|c}
\hline\hline
Method & PSNR & SAM & RMSE & ERGAS & UIQI \\
\hline
Best Values & $+\infty$ & 0 & 0 & 0 & 1 \\
\hline\hline
  Baseline &15.35 &20.41 &0.135 &197.484 &0.457 \\
       CSU &27.73 &3.76 &0.028 &11.215 &0.813 \\
    HySure &27.85 &4.65 &0.033 &12.051 &0.823 \\
      NSSR &25.62 &7.75 &0.051 &11.358 &0.817 \\
      HCTR &27.74 &5.70 &0.039 &11.615 &0.858 \\
    CNNFUS &27.81 &4.91 &0.036 &11.483 &0.857 \\
      MIAE &29.53 &3.59 &0.026 &10.048 &0.875 \\
       SDP &\textbf{32.56} &\textbf{2.61} &\textbf{0.020} &\textbf{9.524} &\textbf{0.959} \\
\hline\hline
\end{tabular}
\end{table}

\begin{figure*}[!t]
\centering
{\includegraphics[width=\myimgsize in]{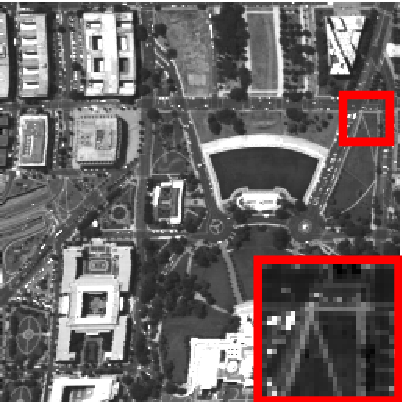}}
{\includegraphics[width=\myimgsize in]{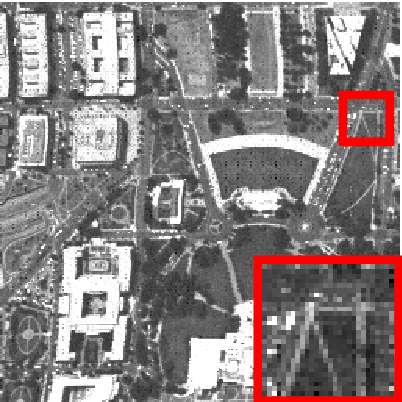}}
{\includegraphics[width=\myimgsize in]{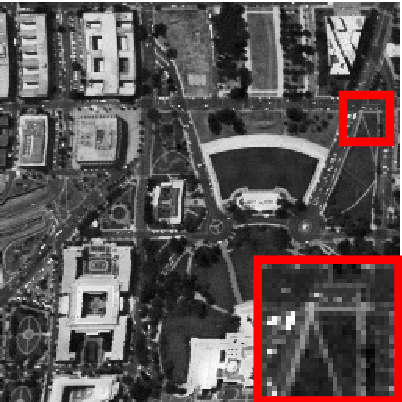}}
{\includegraphics[width=\myimgsize in]{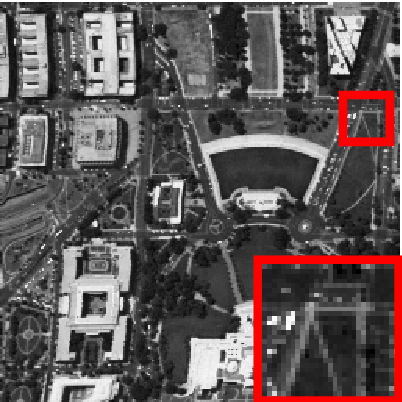}}
{\includegraphics[width=\myimgsize in]{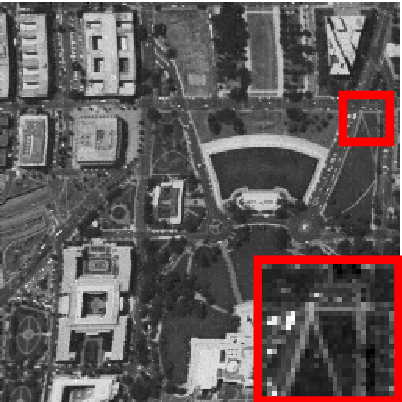}}
{\includegraphics[width=\myimgsize in]{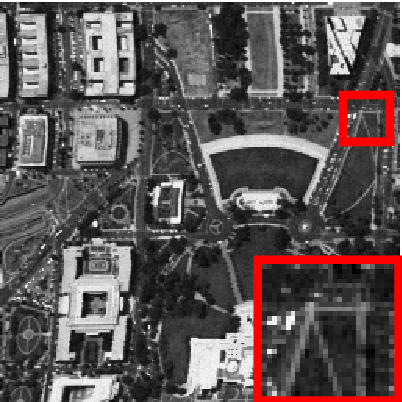}}
{\includegraphics[width=\myimgsize in]{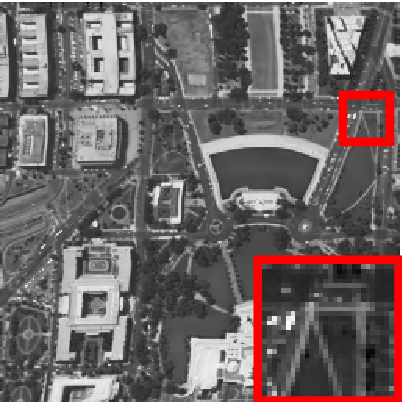}}
{\includegraphics[width=\myimgsize in]{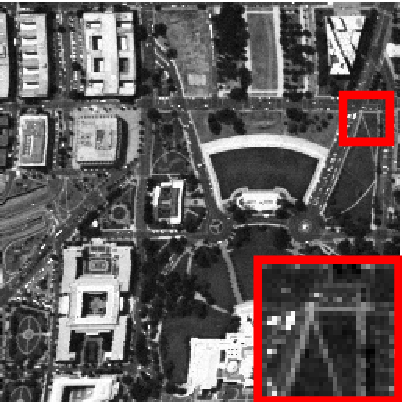}}
{\includegraphics[width=\myimgsize in]{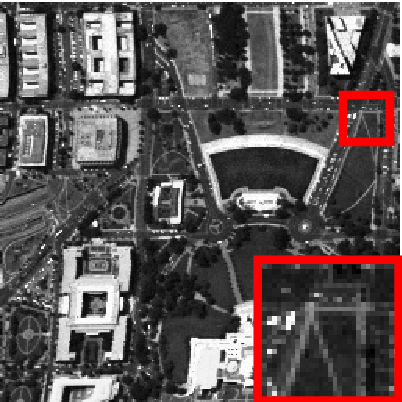}} \\
\subfigure[] {\includegraphics[width=\myimgsize in]{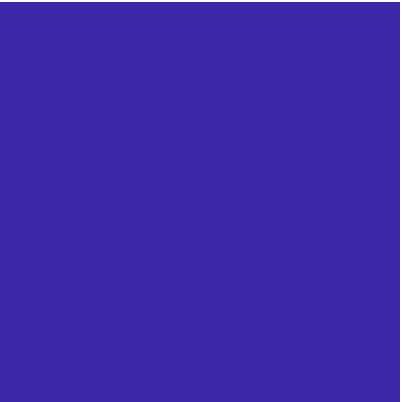}}
\subfigure[] {\includegraphics[width=\myimgsize in]{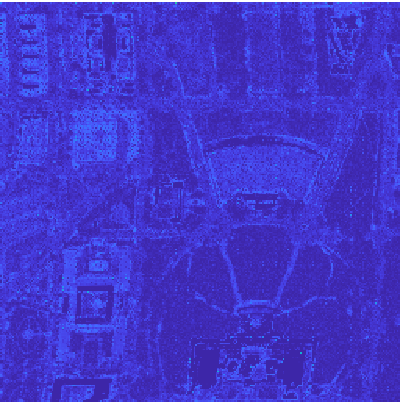}}
\subfigure[] {\includegraphics[width=\myimgsize in]{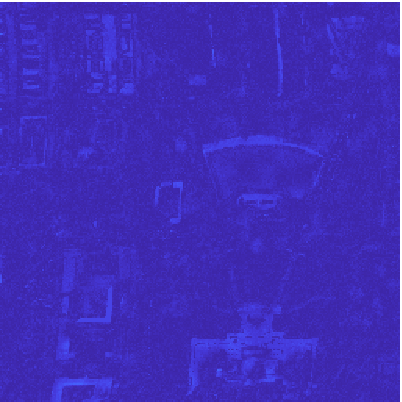}}
\subfigure[] {\includegraphics[width=\myimgsize in]{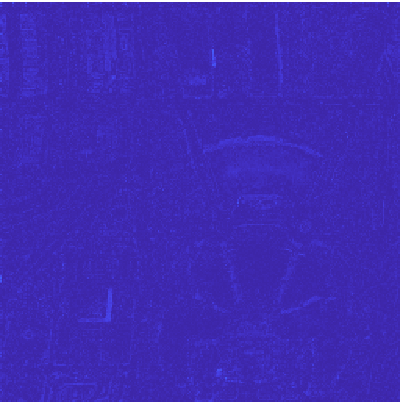}}
\subfigure[] {\includegraphics[width=\myimgsize in]{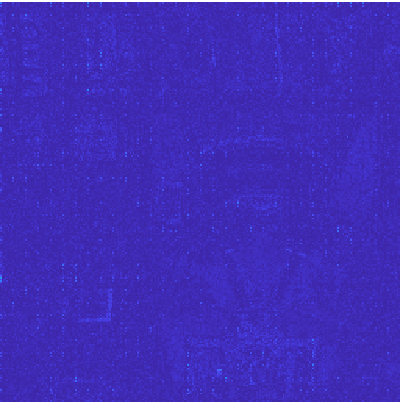}}
\subfigure[] {\includegraphics[width=\myimgsize in]{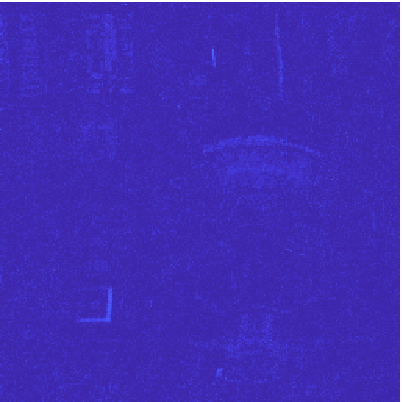}}
\subfigure[] {\includegraphics[width=\myimgsize in]{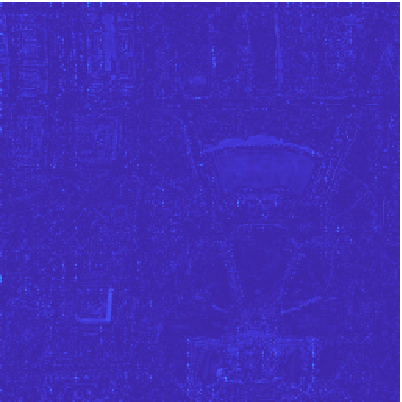}}
\subfigure[] {\includegraphics[width=\myimgsize in]{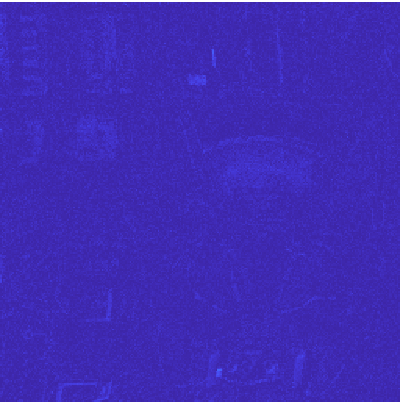}}
\subfigure[] {\includegraphics[width=\myimgsize in]{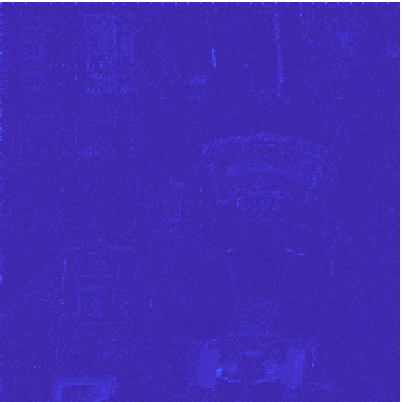}}
\caption{Fused images and error maps at the 30th band when applied to the DC dataset.
(a) Reference image. (b) Baseline. (c) CSU. (d) HySure.
(e) NSSR. (f) HCTR. (g) CNNFUS.
(h) MIAE. (i) SDP.}
\label{fig_dc}
\end{figure*}

\begin{figure*}[!t]
\centering
\subfigure[] {\includegraphics[width=0.3 \textwidth]{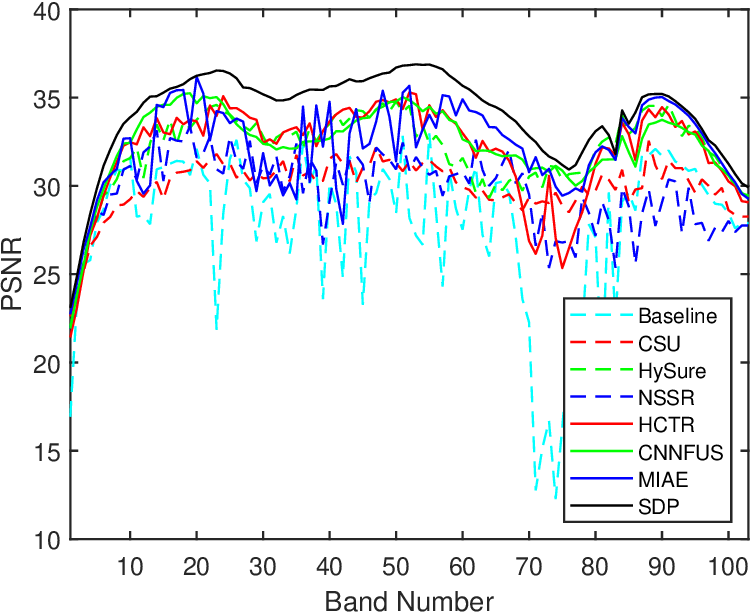}}~~~
\subfigure[] {\includegraphics[width=0.3 \textwidth]{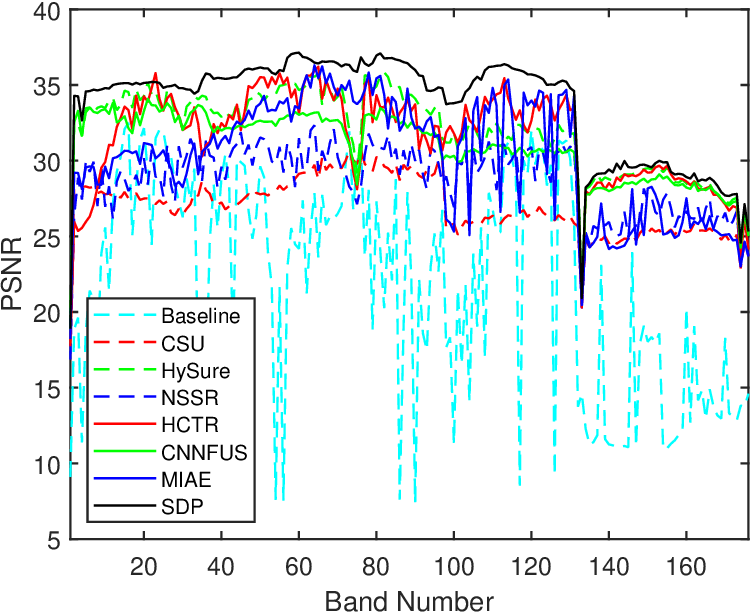}}~~~
\subfigure[] {\includegraphics[width=0.3 \textwidth]{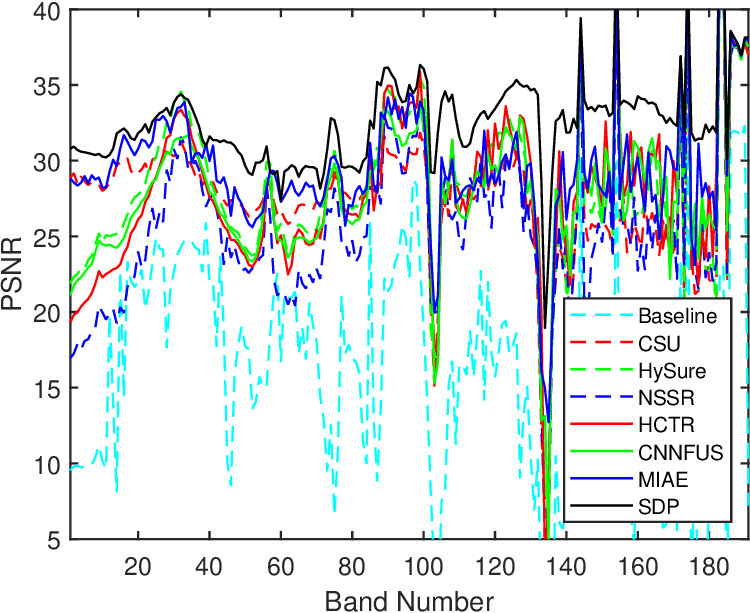}}
\caption{PSNR as a function of spectral band. (a) PaviaU dataset. (b) KSC dataset. (c) DC dataset.}
\label{fig_psnr}
\end{figure*}

\begin{figure*}[!t]
\centering
\subfigure[] {\includegraphics[width=0.3 \textwidth]{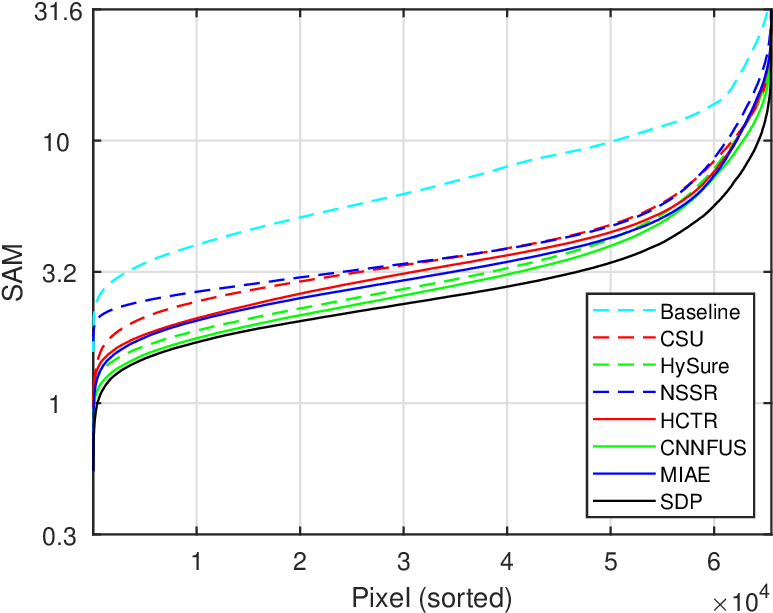}}~~~
\subfigure[] {\includegraphics[width=0.3 \textwidth]{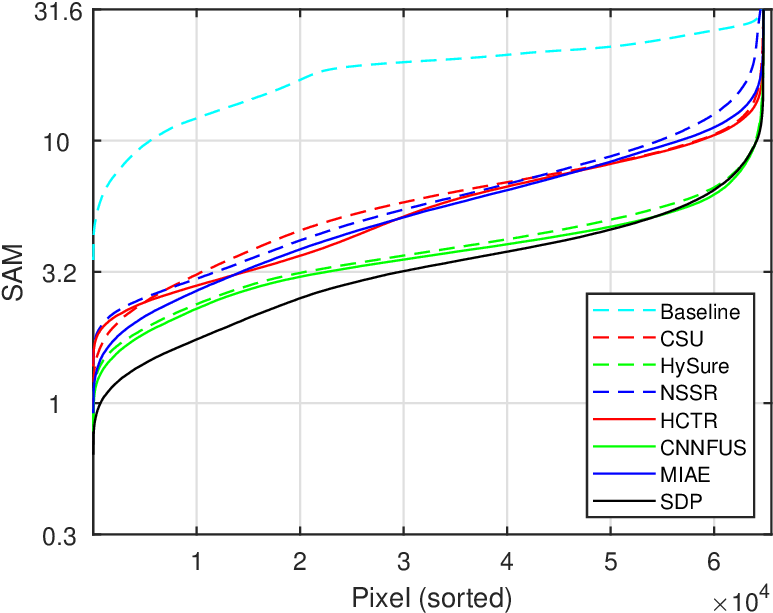}}~~~
\subfigure[] {\includegraphics[width=0.3 \textwidth]{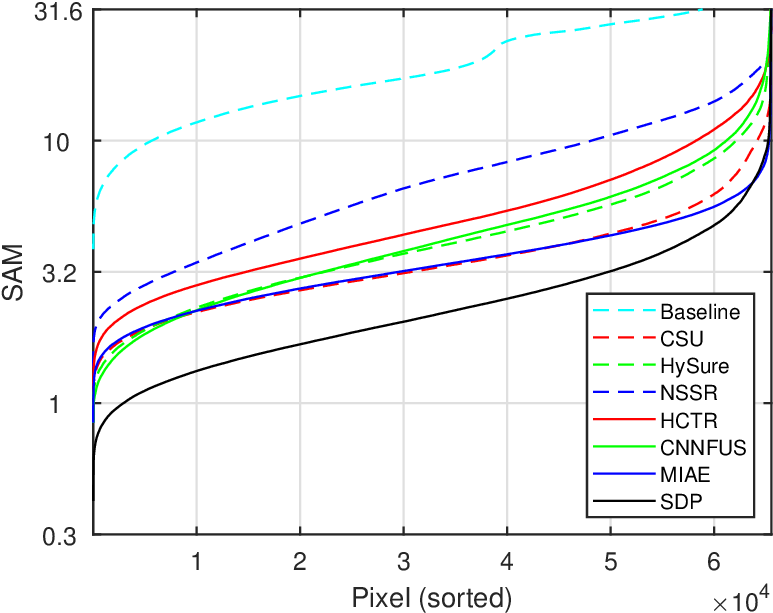}}
\caption{SAM (plotted in a $\log_{10}(\cdot)$ scale) as a function of sorted pixel. (a) PaviaU dataset. (b) KSC dataset. (c) DC dataset.}
\label{fig_sam}
\end{figure*}

This set of experiments is to evaluate the performance of the proposed SDP by comparing with seven related methods. The first method is the baseline one that solves the unconstrained HSI super-resolution problem using the Adam. The baseline method can be seen as the specific version of the proposed SDP, i.e., the problem (\ref{eq_final}) with $\gamma=0$. The next four methods are those that adopt the hand-craft priors, i.e., CSU \cite{lanaras2015hyperspectral}, HySure \cite{simoes2015convex}, NSSR \cite{dong2016hyperspectral} and HCTR \cite{Xu2020Hyperspectral}. The last two methods are based on the specific priors. One is the CNNFUS \cite{dian2021regularizing} method that uses the trained network as a denoiser, and the other is the MIAE \cite{MIAE} method that implements a self-supervised deep prior. The free parameters of these comparison methods are tuned on our datasets. For all comparison methods, they are performed in the blind condition, that is, the PSF ${\bf B}$ and SRF ${\bf R}$ are unknown. The blind estimation network proposed in \cite{MIAE} is adopted to estimate these two functions.

Table \ref{tab_pavia} reports the five quantitative results of the comparison methods on the PaviaU dataset. As seen from this table, all methods show significant improvements when compared to the baseline method. The proposed SDP method performs the best among the comparison methods followed by CNNFUS and MIAE, and its improvements over the second one are near 2 dB in PSNR and $0.5^{\circ}$ in SAM. Fig. \ref{fig_pavia} illustrates the fused images and their error images with respect to the reference image, in terms of the 30th band. The reference image is also included for reference. Visually, it can be seen from the fused images that most methods generate high quality images, while the baseline method has significant spatial distortion. From the error images, it can be observed that SDP and MIAE are closer to the reference image. Fig. \ref{fig_psnr} (a) presents the PSNR of each spectral band for the comparison methods. In this figure, the proposed SDP method gives the best values in almost all bands. Fig. \ref{fig_sam} (a) presents the SAM of each pixel for the comparison methods, with the pixels sorted in order of ascending error. It can be seen that SDP consistently outperforms the other methods at the pixel level followed by CNNFUS.

Table \ref{tab_ksc} presents the PSNR, SAM, RMSE, ERGAS and UIQI results of the comparison methods on the KSC dataset. It can be seen from this table that the baseline method performs the worst, and the improvements of the other methods are over 5 dB in PSNR and $12^{\circ}$ in SAM. The proposed SDP method outperforms the other methods in four metrics. The fused images and their error maps at the 30th band are shown in Fig. \ref{fig_ksc}. From this figure, it can be observed that the baseline method generates the worst results, and the results of SDP are better than the others. To compare the aforementioned methods in different respects, Fig. \ref{fig_psnr} (b) and Fig. \ref{fig_sam} (b) give the PSNR and SAM as functions of spectral band and sorted pixel, respectively. It can be seen that the proposed SDP method can give the best results in most cases.

Table \ref{tab_dc} summarizes the results of the comparison methods under the five quality metrics for the DC dataset. It can be seen from this table that the baseline method gives the worst results, and the proposed SDP method achieves the highest results in all metrics. The improvements of SDP over the second method (MIAE) is about 3 dB in PSNR and $1^{\circ}$ in SAM. Fig. \ref{fig_dc} illustrates the fusion results and error maps of the comparison methods in terms of the 30th band. Through visual inspection, it can be seen that there is obvious spatial distortion in the baseline method, and the results of SDP are closer to the truth. Fig. \ref{fig_psnr} (c) and Fig. \ref{fig_sam} (c) illustrate the PSNR of each spectral band and the SAM of each sorted pixel, respectively. These two figures report that SDP achieves higher results than the others in terms of band-level PSNR and pixel-level SAM.

\subsubsection{Computational Efficiency}
\begin{table*}[!t]
\caption{Running times (in seconds) of the comparison methods}
\label{tab_time}
\centering
\begin{tabular}
{c|c|c|c|c|c|c|c|c}
\hline\hline
Method & Baseline & CSU & HySure & NSSR & HCTR & CNNFUS & MIAE & SDP \\
\hline\hline
PaviaU &6.3 &38.1 &45.1 &26.2 &17.9 &1.5 &39.3 &119.7 \\
KSC &10.9 &12.6 &44.9 &40.5 &27.2 &2.0 &57.9 &145.1 \\
DC &12.1 &23.4 &45.0 &44.9 &30.6 &1.9 &36.2 &150.4 \\
\hline\hline
\end{tabular}
\end{table*}

The experiments are carried out using a desktop computer with an Intel Core i9-13900K CPU, a GeForce RTX 4090 GPU, and 64 GB memory. There are three methods that are implemented by the PyTorch framework, i.e., Baseline, MIAE and SDP. The remaining methods CSU, HySure, NSSR, HCTR and CNNFUS are performed using MATLAB. Table \ref{tab_time} summarizes the running
times of these comparison methods for reference.

\subsection{Experiment Results on Real Data}
\begin{figure*}[!t]
\centering
\subfigure[] {\includegraphics[width=\myimgsizer in]{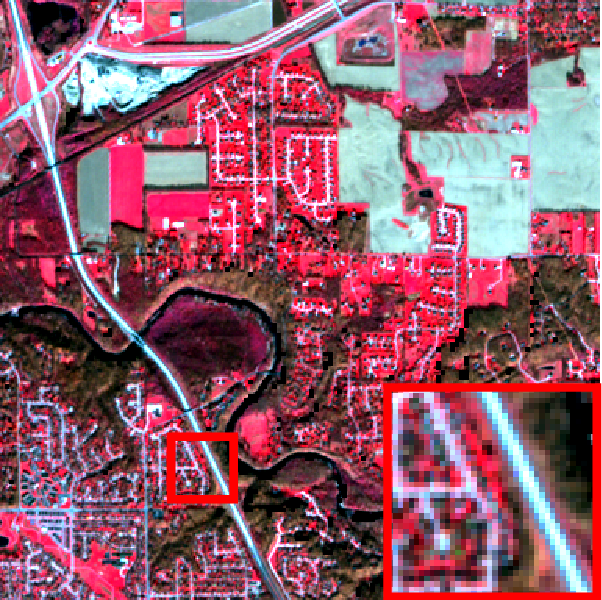}}
\subfigure[] {\includegraphics[width=\myimgsizer in]{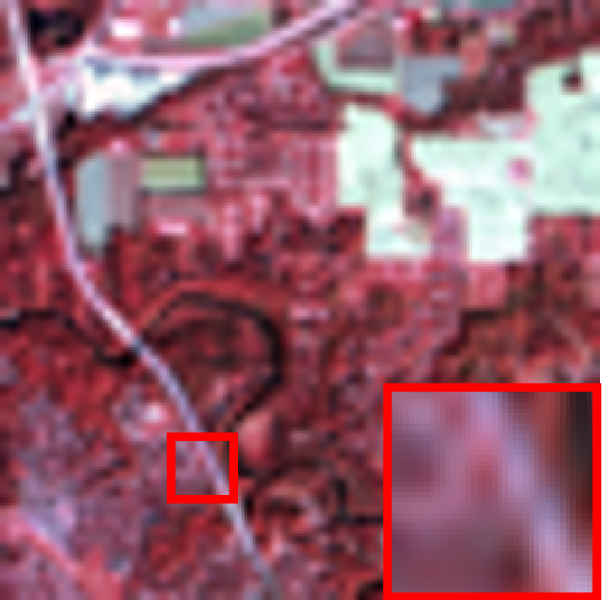}}
\subfigure[] {\includegraphics[width=\myimgsizer in]{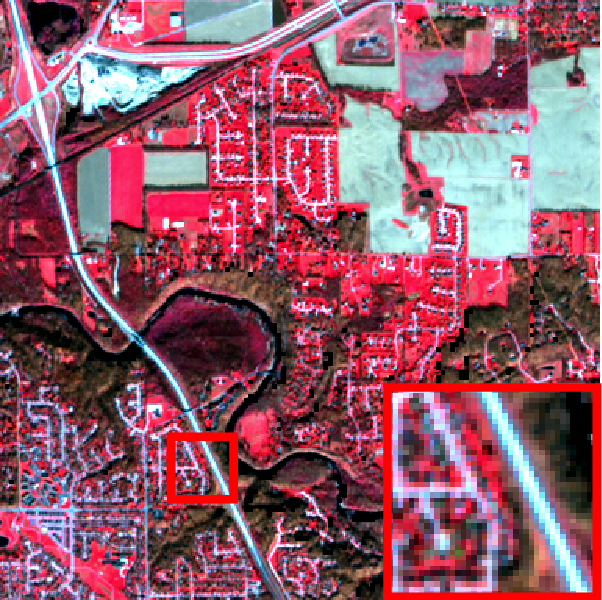}}
\subfigure[] {\includegraphics[width=\myimgsizer in]{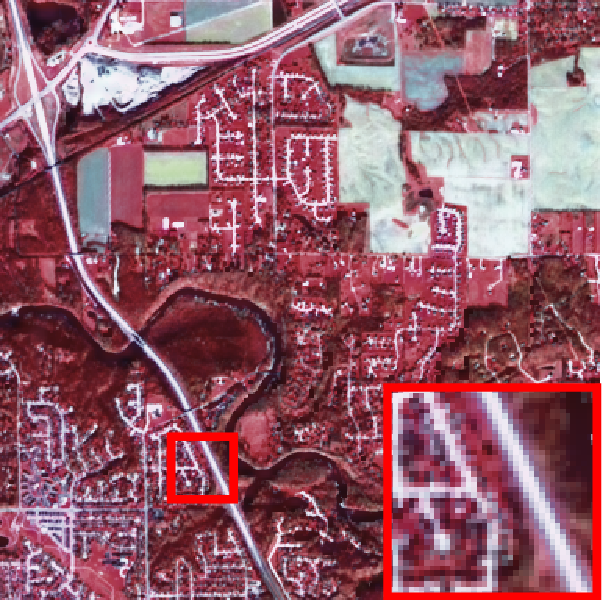}}
\subfigure[] {\includegraphics[width=\myimgsizer in]{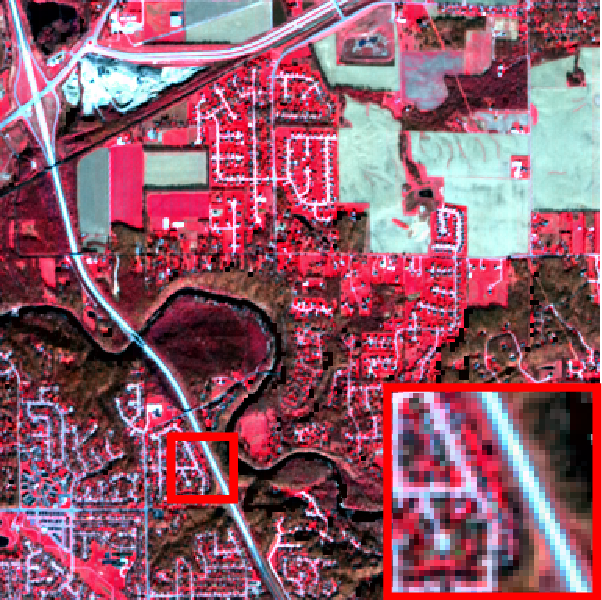}} \\
\subfigure[] {\includegraphics[width=\myimgsizer in]{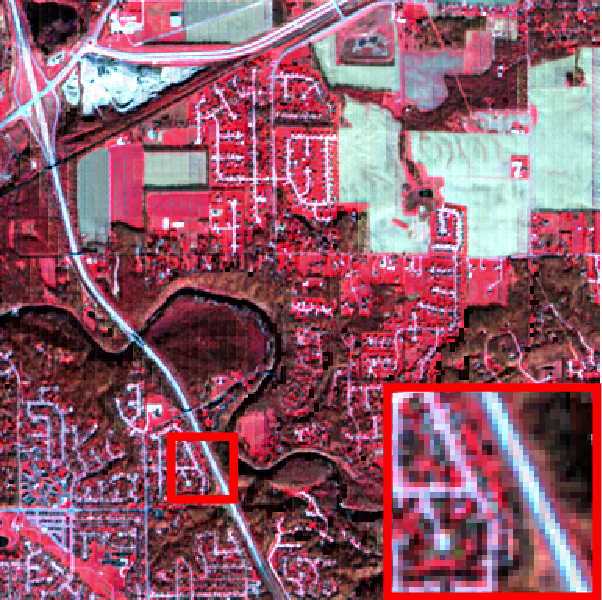}}
\subfigure[] {\includegraphics[width=\myimgsizer in]{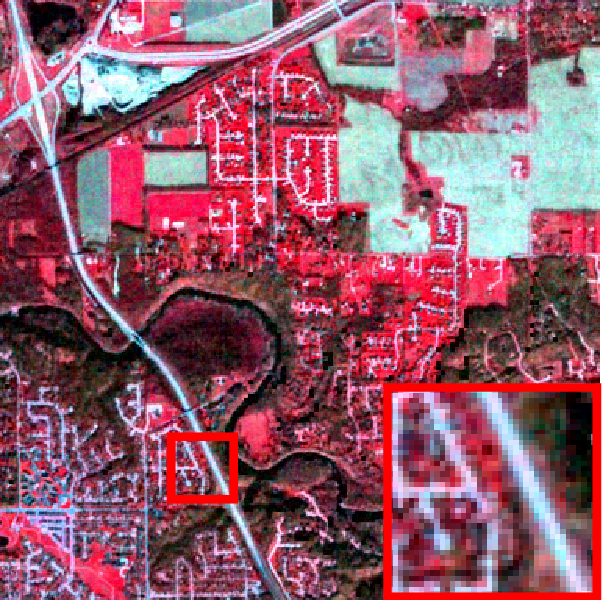}}
\subfigure[] {\includegraphics[width=\myimgsizer in]{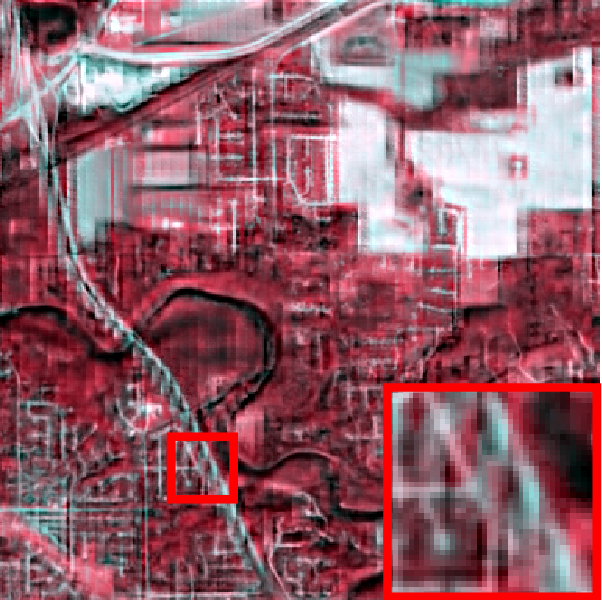}}
\subfigure[] {\includegraphics[width=\myimgsizer in]{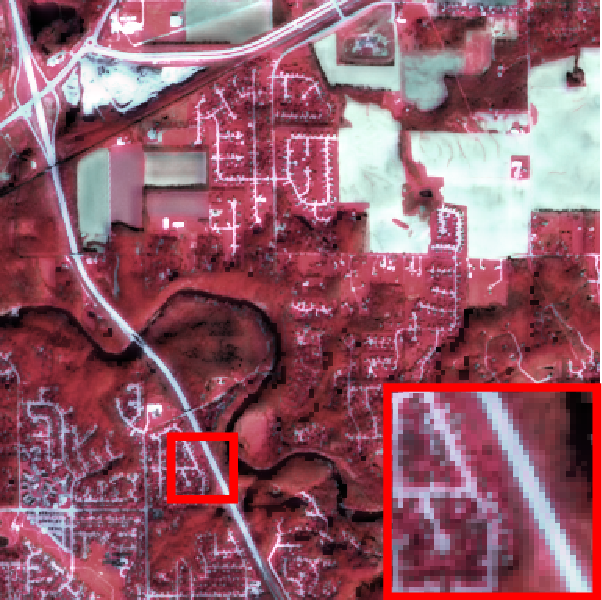}}
\subfigure[] {\includegraphics[width=\myimgsizer in]{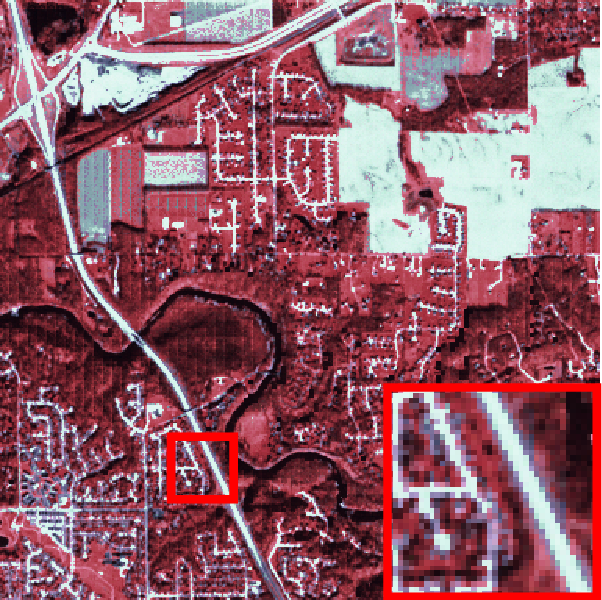}}
\caption{False color images of fusion results when applied to the real Hyperion-Sentinel dataset.
(a) HR-MSI. (b) LR-HSI. (c) Baseline.
(d) CSU. (e) HySure. (f) NSSR.
(g) HCTR. (h) CNNFUS. (i) MIAE. (j) SDP.}
\label{fig_real}
\end{figure*}

\begin{table*}[!t]
\caption{Quality metrics without reference for the real Hyperion-Sentinel dataset (the best values are marked in bold)}
\label{tab_real}
\centering
\begin{tabular}
{c|c||c|c|c|c|c|c|c|c}
\hline\hline
 & Best & Baseline & CSU & HySure & NSSR & HCTR & CNNFUS & MIAE & SDP \\
\hline\hline
$D_{\lambda}$ & 0 &0.1551 &0.0518 &0.1271 &0.1353 &0.0598 &0.0567 &0.0409 &\textbf{0.0228} \\
     $D_s$ & 0 &0.2718 &0.1136 &0.0975 &0.1472 &\textbf{0.0422} &0.1136 &0.0708 &0.0704 \\
       QNR & 1 &0.6152 &0.8405 &0.7878 &0.7374 &0.9005 &0.8362 &0.8912 &\textbf{0.9084} \\
\hline\hline
\end{tabular}
\end{table*}

A real spaceborne dataset \cite{TBCNN} is used to evaluate SDP in practical application. The LR-HSI is acquired by Hyperion sensor carried on the Earth Observing-1 satellite, and the HR-MSI is acquired by the Sentinel-2A satellite. These two images are taken over Lafayette, LA, USA in October and November, 2015, respectively. The Hyperion sensor is characterized by 242 spectral bands, with a spectral range of 0.4 to 2.5 $\mu m$. The noisy bands and water absorption bands are removed, leaving 83 bands. The spatial resolution of LR-HSI is 30 m. The HR-MSI has 13 bands, where four bands with 10 m spatial resolution are kept. The LR-HSI is spatially downsampled with a factor of 2 in two directions. Then, we crop a $64 \times 64$-pixel HSI and a $384 \times 384$-pixel MSI as observation images from the overlapped region of the LR-HSI and HR-MSI, respectively. The LR-HSI is used as the clean hyperspectral pixels. Table \ref{tab_real} reports the results of three metrics without reference for the comparison methods mentioned in Section \ref{sec_syn}. It can be seen that the proposed SDP gives the highest QNR and $D_{\lambda}$, and HCTR gives the highest $D_{s}$.  Fig. \ref{fig_real} illustrates the false color images of the observation images and the fusion results of the comparison methods. Visually, it can be observed that SDP shows good performance in terms of color and brightness.

\section{Conclusion}
\label{sec_con}
This paper has proposed a SDP method for fusion-based HSI super-resolution. The proposed SDP belongs to the model-based approaches, where a spectral diffusion prior is exploited to enforce the desired solution. Firstly, a spectral diffusion model is proposed to model the spectral data distribution of HSIs and thereby to generate high-quality spectra. Unlike the methods of other data modalities, such as image and video, the spectral diffusion model does not simply extend the existing diffusion models, but rather depicts the spectral diffusion and generative processes. Accordingly, a simple and effective MLP-based denoising network is adopted rather than the commonly used U-Net denoising network. Secondly, in the framework of maximum a posteriori, a unified optimization model with the spectral diffusion model as a prior is proposed to perform the fusion of LR-HSI and HR-MSI by assuming the target HR-HSI follow a given spectral data distribution. To achieve this, the transition information of every two neighboring Markov states is kept as the constraints in the inverse generative process, and the constraints are further transformed into the tractable denoising loss under KL divergence. Finally, the objective optimization problem is solved by the Adam with the solution as trainable parameters, and its subproblems with respect to the timesteps are solved sequentially according to the reverse generative process. The experimental results conducted on three synthetic datasets and one real dataset demonstrate the effectiveness of the proposed SDP, and show that FID is suitable to measure the generation quality of spectra and can be used to guide the design of spectral diffusion model to improve the fusion quality. Although the results obtained by SDP are very encouraging, further enhancement such as high-quality spectral generative model and high-performance  algorithms should be pursued in future.


\section*{Acknowledgment}
The authors would like to thank the authors of \cite{lanaras2015hyperspectral,simoes2015convex,dong2016hyperspectral,Xu2020Hyperspectral,dian2021regularizing} for providing their codes.

\ifCLASSOPTIONcaptionsoff
  \newpage
\fi

\bibliographystyle{IEEEtran}
\bibliography{IEEEabrv,SDP}

\end{document}